%% file: main.tex
\documentclass{article} 
\usepackage{iclr2026_conference,times}

\input{math_commands.tex}

\usepackage{hyperref}
\usepackage{url}
\usepackage{graphicx}
\usepackage[utf8]{inputenc}
\usepackage[T1]{fontenc}
\usepackage{times}
\usepackage{hyperref}
\usepackage{amsmath}
\usepackage{url}
\usepackage{booktabs}
\usepackage{multirow}
\usepackage{natbib} 
\usepackage{comment}
\usepackage{subcaption}
\usepackage{graphicx}
\usepackage{booktabs}
\usepackage{pifont}
\usepackage{tabularx}
\usepackage{booktabs}
\usepackage{siunitx}
\usepackage{pdfpages}
\usepackage{svg}              
\svgsetup{inkscapelatex=true}

\usepackage{listings}
\usepackage{xcolor}

\definecolor{codegreen}{rgb}{0,0.6,0}
\definecolor{codegray}{rgb}{0.5,0.5,0.5}
\definecolor{codepurple}{rgb}{0.58,0,0.82}
\definecolor{backcolour}{rgb}{0.97,0.97,0.97}
\definecolor{stringcolor}{rgb}{0.75,0.25,0.25}

\lstdefinestyle{pystyle}{
    language=Python,
    backgroundcolor=\color{backcolour},   
    commentstyle=\color{codegreen}\itshape,
    keywordstyle=\color{blue}\bfseries,
    numberstyle=\tiny\color{codegray},
    stringstyle=\color{stringcolor},
    basicstyle=\ttfamily\small,
    breakatwhitespace=false,         
    breaklines=true,                 
    captionpos=b,                    
    keepspaces=true,                                    
    showspaces=false,                
    showstringspaces=false,
    showtabs=false,                  
    tabsize=4,
    frame=single,
    rulecolor=\color{black!30},
    xleftmargin=2em,
    xrightmargin=2em,
    aboveskip=1.5em,
    belowskip=1em,
    numbers=none  
}

\lstdefinelanguage{Markdown}{
  sensitive=false,
  morecomment=[s]{<!--}{-->},
  morestring=[b]`,
  literate=
    {*}{{{\color{blue}*}}}1
    {-}{{{\color{blue}-}}}1
    {+}{{{\color{blue}+}}}1
    {_}{{{\color{blue}\_}}}1
    {~}{{{\color{blue}\~{}}}}1
    {[}{{{\color{codegreen}[}}}1
    {]}{{{\color{codegreen}]}}}1
    {(}{{{\color{codegreen}(}}}1
    {)}{{{\color{codegreen})}}}1
    {!}{{{\color{codegreen}!}}}1
}

\lstdefinestyle{mdstyle}{
    language=Markdown,
    backgroundcolor=\color{backcolour},
    commentstyle=\color{codegray}\itshape,   
    keywordstyle=\color{blue}\bfseries,      
    numberstyle=\tiny\color{codegray},
    stringstyle=\color{stringcolor},        
    basicstyle=\ttfamily\small,
    breakatwhitespace=false,
    breaklines=true,
    captionpos=b,
    keepspaces=true,
    showspaces=false,
    showstringspaces=false,
    showtabs=false,
    tabsize=4,
    frame=single,
    rulecolor=\color{black!30},
    xleftmargin=2em,
    xrightmargin=2em,
    aboveskip=1.5em,
    belowskip=1em,
    numbers=none
}

\definecolor{backcolour}{RGB}{245,245,248}
\definecolor{codegreen}{RGB}{34,139,34}
\definecolor{codegray}{RGB}{120,120,120}
\definecolor{stringcolor}{RGB}{163,21,21}

\usepackage[most]{tcolorbox}
\usepackage{xcolor}

\title{MCP4IFC: IFC-Based Building Design using Large Language Models}

\author{Bharathi Kannan Nithyanantham$^{1}$ \hskip1em 
Tobias Sesterhenn$^{2}$ \hskip1.5em 
Ashwin Nedungadi$^{1}$ \hskip1.5em
\\\textbf{Sergio Peral Garijo$^{2}$ \hskip1.5em 
Janis Zenkner$^{2}$ \hskip1.5em 
Christian Bartelt$^{2}$ \hskip1.5em 
Stefan Lüdtke$^{1}$ }
\\$^1$University of Rostock \hskip2em 
$^2$Technical University of Clausthal
\\\texttt{bharathikannan.nithyanantham@uni-rostock.de} 
\\\texttt{tobias.sesterhenn@tu-clausthal.de}
}

\date{October 2025}

\usepackage[table]{xcolor}
\newcolumntype{Y}{>{\raggedright\arraybackslash}X}

\newcommand{\cmark}{\textcolor{green!60!black}{\checkmark}}
\newcommand{\xmark}{\textcolor{red!70!black}{\texttimes}}
\newcommand\C[1]\null

\iclrfinalcopy 
\begin{document}
\maketitle

\maketitle

\begin{abstract}
Bringing generative AI into the architecture, engineering and construction (AEC) field requires systems that can translate natural language instructions into actions on standardized data models. We present MCP4IFC, a comprehensive open-source framework that enables Large Language Models (LLMs) to directly manipulate Industry Foundation Classes (IFC) data through the Model Context Protocol (MCP). The framework provides a set of BIM tools, including scene querying tools for information retrieval, predefined functions for creating and modifying common building elements, and a dynamic code-generation system that combines in-context learning with retrieval-augmented generation (RAG) to handle tasks beyond the predefined toolset. Experiments demonstrate that an LLM using our framework can successfully perform complex tasks, from building a simple house to querying and editing existing IFC data. Our framework is released as open-source to encourage research in LLM-driven BIM design and provide a foundation for AI-assisted modeling workflows.
Our code is available at \url{https://show2instruct.github.io/mcp4ifc/}
\end{abstract}






\section{Introduction}

Building Information Modeling (BIM) has transformed how the architecture, engineering and construction (AEC) fields work. Instead of using separate 2D drawings, teams share a single 3D model that contains all the important details about a building \citep{SUCCAR2009357}. This enables seamless collaboration and more efficient coordination throughout the project lifecycle. These models can include not only building meshes, but also information about materials, costs, schedules, and more.  The Industry Foundation Classes (IFC) standard defines a common schema for storing and sharing BIM data, so different software platforms can use the same models \citep{ISO16739-1_2024}. However, creating, editing, and maintaining BIM models remains complex and is typically handled by experts. Meanwhile, rapid progress in Large Language Models (LLMs) suggests a new paradigm for interacting with complex software. Modern LLMs can interpret natural language instructions and act as intelligent agents that plan and execute tool calls to achieve user goals \citep{wang2023voyager, yao2023react}. This presents an exciting possibility in the BIM context: an AI assistant that enables architects and engineers to describe desired changes or questions in natural language and have the model automatically execute corresponding BIM operations.

Recent studies have begun exploring natural language interfaces for BIM, focusing on either model generation and editing (e.g., \cite{du2024text2bim, deng2025bimgent}) or on information querying (e.g., \cite{Hellin2025Natural, li2025interactive}). However, most existing systems rely on high-level APIs from proprietary software such as Revit \citep{AutodeskRevit_Overview} or Vectorworks \citep{Vectorworks_OfficialSite}. These APIs expose operations at the application layer with their own schema and do not interact directly with the IFC data structure. 

\begin{figure}[t]
  \centering
  \includegraphics[width=\linewidth]{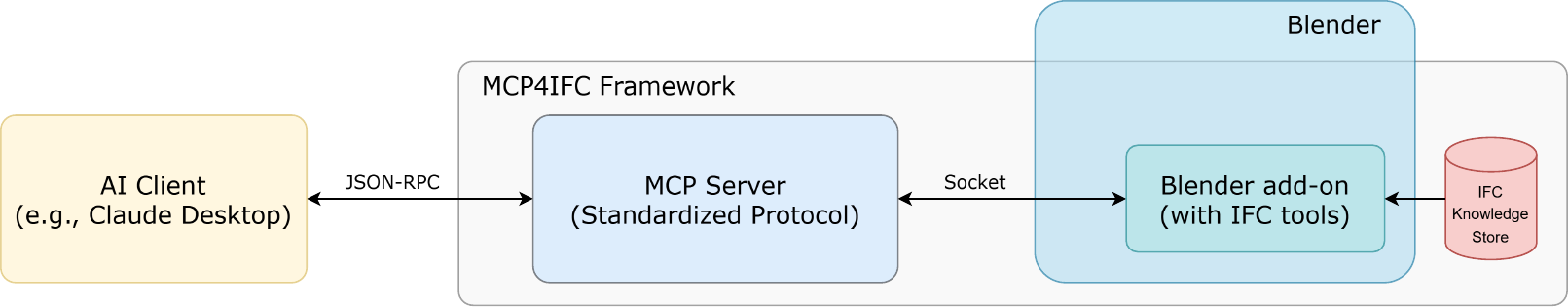}
  \caption{Overall system architecture of our MCP4IFC framework}
  \label{overview_image}
\end{figure}

To address this limitation, we present MCP4IFC, a framework that leverages the Model Context Protocol (MCP) \citep{ModelContextProtocol_Introduction} to enable LLMs to read, create, and edit IFC models directly via high-level tool calls. The overall system architecture is illustrated in Figure \ref{overview_image}. The framework operates as a local MCP server and utilizes the open-source IfcOpenShell library \citep{ifcopenshell} together with the Bonsai add-on \citep{BonsaiBIM_homepage} in Blender. The MCP is an open standard that provides a single protocol that any LLM client can use \citep{hou2025model}. Some projects have already explored connecting MCP with 3D modeling environments. For instance, the Blender MCP integration \citep{ahujasid_blender-mcp} generates Blender API code through a code execution tool, and a Bonsai MCP plugin \citep{JotaDeRodriguez_Bonsai_mcp} exposing a few BIM operations to query IFC models and place objects. Our framework introduces a complete and standardized solution that uses IfcOpenShell as the main toolkit for working with IFC data and supports full query, create, and edit operations. All functionalities are implemented via IfcOpenShell API calls, with Bonsai acting only as a wrapper to interact with Blender. As all components are open source, we are not constrained by closed APIs or automation limits of commercial BIM software. This modular design makes it straightforward to adapt the tools to proprietary environments.

A key innovation of our framework is combining In-Context Learning (ICL) \citep{10.5555/3495724.3495883} with Retrieval-Augmented Generation (RAG) \citep{10.5555/3495724.3496517} to dynamically generate Python code for IFC-specific tasks. ICL refers to the model's ability to learn new tasks or patterns during inference by simply showing examples in the prompt. ICL is the core of our framework, which guides the model to generate a specific tool by providing detailed steps. This allows the model to go beyond the provided fixed toolset. When the model's prior knowledge is insufficient (e.g., with the IfcOpenShell API), the LLM can invoke an optional RAG tool. This module retrieves relevant documentation and code examples from a local vector store and injects them into the prompt. The retrieved information provides additional context for ICL to generate accurate code. This design follows the retrieval-augmented code generation approach, where LLMs search large codebases or external documentation to produce accurate and context-aware code \citep{Zhou2022DocPromptingGC, 10.5555/3737916.3739517, Wang2024CodeRAGBenchCR}.

We demonstrate that LLMs can design and modify IFC models by following step-by-step natural-language instructions. For example, an LLM can incrementally construct a small house using only standardized tool calls and achieve scene-query performance competitive with state-of-the-art methods \citep{Hellin2025Natural}. These indicate that standardized tool use enables LLMs to work effectively across the BIM lifecycle. To our knowledge, our framework is the first comprehensive MCP server that allows LLMs to create, edit, and query IFC models while also supporting dynamic tool generation.

Our main contributions are:
\begin{enumerate}
    \item \textbf{An open-source MCP framework for IFC}: Our framework offers a comprehensive set of more than 50 BIM tools, enabling LLMs to create, modify, and query IFC models through standardized MCP tool calls.
    \item \textbf{Dynamic tool expansion through ICL and RAG}: Our framework combines ICL with a RAG system to support on-the-fly code generation for IFC-specific tasks beyond a predefined toolset.
    \item \textbf{End-to-end BIM manipulation}: We show that existing multimodal LLMs can successfully perform IFC querying, editing, and generation tasks using only natural language and tool calls, without direct user interaction with BIM software.
\end{enumerate}

\section{Related Work}

Applying LLMs in the AEC field is an active, rapidly growing area of research. Our work focuses on tool-augmented LLMs, where models can invoke external tools to create and modify IFC-based BIM data. This section reviews related studies on LLMs for tool use, their application to BIM and 3D design workflows, and their ability to understand spatial information. 

\subsection{LLMs as Tool-Using Agents}
Augmenting LLMs with external tools has become an active area of research in recent years. This line of work focuses on enabling models to take actions in the external world via tool calls. ReAct \citep{yao2023react} introduced a framework where an LLM iteratively plans actions, calls external tools, observes the results, and then uses that feedback to decide the next step. Toolformer \citep{10.5555/3666122.3669119} showed that LLMs can learn when and how to use APIs without requiring task-specific labels. Approaches such as Gorilla \citep{Patil2023GorillaLL} train LLMs on large, diverse API specifications and use retrieval to select the appropriate function. Datasets like ToolBench \citep{qin2023toolllm} provide thousands of tool-use instructions for model training. As tool-use capabilities become more common, major LLM providers expose APIs that allow models to call external functions or tools directly. However, these interfaces are not compatible across providers, leading to integration and scalability challenges. The  MCP \citep{ModelContextProtocol_Introduction} addresses this issue by defining a standardized interface for tool discovery and invocation of these tools over HTTP or local IPC. MCP consists of two main components: an AI Client (typically the LLM application) and an MCP Server, which exposes tools, along with their descriptions and input/output formats using JSON Schema. Because the protocol is standardized, any AI Client can connect to the MCP Server without requiring custom integration code. In our work, we adopt MCP for IFC-based BIM tasks. 


\subsection{LLMs for BIM and 3D Design}
Recent work has explored how LLMs can assist in the design of buildings and 3D structures.
This refers to tasks such as creating and modifying elements, and retrieving information from models for question answering or more advanced tasks, such as clash detection. Most systems are designed to work on either generation/editing \citep{deng2025bimgent, jang2024automated, duggempudi2025text, du2024towards, fernandes2024gpt, yuan20243d} or information retrieval \citep{zheng2023dynamic, Hellin2025Natural, li2025interactive, guoarcbim, liu2025bimcoder}.
We refer to Table \ref{tab:llm_bim} for a categorical overview.

\paragraph{BIM Creation and Modification}
Several systems let LLMs generate code that interacts directly with proprietary BIM software APIs. Text2BIM \citep{du2024text2bim} and previous work \citep{du2024towards} employ multiple LLM agents to call custom tools built on the Vectorworks API to construct BIM models from textual descriptions. NADIA \citep{jang2024automated} and Text-to-Layout \citep{duggempudi2025text} translate user instructions into intermediate JSON actions (e.g., create, modify, retrieve, delete) that are then converted into executable Revit API calls. \cite{dong2025ai} uses the AutoGen framework \citep{wu2024autogen} with the Revit API to coordinate specialized agents for wall generation, clash detection, and information retrieval, while \cite{fernandes2024gpt} takes a simpler approach by representing BIM data as CSV files and exposing a fixed set of operations (e.g., hiding or coloring elements).
Although these systems enable fine-grained model manipulation, they remain tightly coupled to closed-source platforms, limiting portability and extensibility.
A different direction extends beyond APIs by simulating user interactions through the software’s graphical interface. BIMgent \citep{deng2025bimgent} enables an LLM to control BIM software via mouse clicks and keystrokes, using screenshots of the current model state as visual feedback for iterative refinement. Similarly, 3D-PreMise \citep{yuan20243d} integrates rendered images of generated 3D shapes to help the model construct and improve them within Blender.
Inspired by these works, our system also incorporates tools for visual feedback, allowing the LLM to assess and respond to the current state of model generation.

\begin{table*}[tbhp]
\centering
\footnotesize
\renewcommand{\arraystretch}{1.25}
\setlength{\tabcolsep}{5pt}
{
\rowcolors{2}{gray!5}{white}
\begin{tabularx}{\textwidth}{
l c c c Y Y Y Y
}
\toprule
\textbf{Paper} & \textbf{Gen.} & \textbf{Edit.} & \textbf{Retr.} &
\textbf{Framework} & \textbf{Input} & \textbf{Output} & \textbf{Platform} \\
\midrule

\cite{du2024text2bim} & \cmark & \cmark & (\cmark)$^{*}$ & Multi-agent & Text & API calls (Python DSL) & Vectorworks \\
\cite{deng2025bimgent} & \cmark & \cmark & \xmark & Multi-agent & Text + Image & GUI DSL & Tool-agnostic \\
\cite{jang2024automated} & (\cmark)$^{*}$ & \cmark & \xmark & Single-agent & Text & JSON $\rightarrow$ Revit API & Revit \\
\cite{duggempudi2025text} & \cmark & \xmark & \xmark & Single-agent & Text + JSON & API calls (via intermediate JSON) & Revit \\
\cite{du2024towards} & \cmark & \cmark & \xmark & Single-agent + RAG & Text & API calls (Python DSL) & Vectorworks \\
\cite{dong2025ai} & \cmark & \cmark & \cmark & Multi-agent & Text & API calls & Revit \\
\cite{fernandes2024gpt} & \xmark & \cmark & \xmark & Single-agent & Text + CSV & API calls (Python DSL) & Revit \\
\cite{yuan20243d} & \cmark & \cmark & \xmark & Single-agent & Text + Image & Python Code & Blender \\
\cite{zheng2023dynamic} & \xmark & \xmark & \cmark & Single-agent & Text & MongoDB query & Tool-agnostic \\
\cite{Hellin2025Natural} & \xmark & \xmark & \cmark & Single-agent & Text & IfcOpenShell calls (Python DSL) & Tool-agnostic \\
\cite{li2025interactive} & \xmark & \xmark & \cmark & Multi-agent & Text & Spatial Reasoning Calls (Python DSL) & Tool-agnostic \\
\cite{guoarcbim} & \xmark & \xmark & \cmark & Multi-agent & Text & DSL code & Tool-agnostic \\
\cite{liu2025bimcoder} & \xmark & \xmark & \cmark & Single-agent & Text & IFCQL query code & Tool-agnostic \\
\hline
MCP4IFC (ours) & \cmark & \cmark & \cmark & Single-agent + RAG & Text + Image & Tool calls (IfcOpenShell) & mostly Tool-agnostic \\
\bottomrule
\end{tabularx}
}
\caption{
\textbf{Comparison of LLM-based frameworks for BIM and 3D design tasks.}
Checkboxes (\cmark, \xmark) indicate whether the system supports BIM generation (Gen.), editing (Edit.), or information retrieval (Retr.).
$^{*}$ indicates partial or limited support.
}
\label{tab:llm_bim}
\end{table*}

\paragraph{BIM Scene Querying}

Beyond generative workflows, LLMs have also been employed to retrieve information from existing BIM models. For instance, \cite{zheng2023dynamic} builds a framework that preprocesses data into a MongoDB database, enabling an LLM to retrieve information by iteratively translating natural-language prompts into database queries. \cite{Hellin2025Natural} performs agentic information retrieval, using a fixed library of functions to retrieve data directly from an IFC file. \cite{li2025interactive} integrates a tree-based spatial indexing method with a multi-agent LLM framework, enabling the decomposition of spatial queries into geometric reasoning subtasks for efficient retrieval of 3D spatial relationships. \cite{guoarcbim} presents ARCBIM, which uses multiple specialized LLMs and an alignment-refinement coder to convert natural language into executable code for BIM databases. Similarly, BIMCoder \citep{liu2025bimcoder} translates natural language queries into SQL-like or IFCQL statements. 

The main limitation of existing generative systems is their tight coupling to a specific platform, such as Revit \citep{jang2024automated, duggempudi2025text, dong2025ai, fernandes2024gpt} or Vectorworks \citep{du2024text2bim, du2024towards}. This approach lacks portability and requires significant development to support other platforms. Our MCP4IFC framework addresses this challenge by building directly on the IFC standard \citep{ISO16739-1_2024}. By operating on this neutral, open-standard data representation, our server can be easily extended to other platforms that support IFC. This allows a single set of tools to be developed once and used anywhere in the BIM ecosystem. 

\subsection{LLMs for Spatial Understanding}
Working with BIM data requires models to understand complex 3D spatial structures, including geometry, topology, and the relationships between building components. However, while language models excel at sequential text processing, they lack a native grounding in geometry, topology, and spatial relationships. To bridge this gap, current research has focused on vision-language grounding, connecting textual inputs to structured and unstructured 3D data for end-to-end spatial reasoning \citep{li2025zeroshot3dvisualgrounding, 3DGRAND, sceneverse_2024, Zhang2025FromFT}. A major line of work leverages structured representations, such as 3D scene graphs, to encode object semantics, functions, and spatial relations. Advances include open-vocabulary relational reasoning \citep{zhang2025openvocabularyfunctional3dscene}, affordance detection \citep{engelbracht2024spotlightroboticsceneunderstanding}, and enrichment with LLM common-sense knowledge \citep{chen2022extractingzeroshotcommonsense}. In AEC contexts, real-time integration between BIM and scene graphs is being explored \citep{longo2025pixelstographrealtimeintegrationbuilding, schulz2023towards}. Nevertheless, these graphs struggle with geometric fidelity, scalability to large BIM models, and dynamic updates, limiting their utility for tasks such as clash detection or compliance checking.
To address these limitations, other works adopt implicit neural representations, such as NeRFs \citep{mildenhall2020nerfrepresentingscenesneural}, capturing continuous geometry at higher resolution. Hybrid approaches are emerging that embed spatial knowledge directly into the LLM, through spatial tokenization or 3D-informed architectures \citep{mao2025spatiallmtraininglargelanguage, ma2025spatialllmcompound3dinformeddesign}, or by projecting point-cloud features into the language embedding space for unified training \citep{hong20233dllminjecting3dworld}. These shift 3D information from an external structure into a first-class element of the generative reasoning process.
A complementary direction uses LLMs for text-to-simulation, generating executable scene construction code (e.g., Blender or Unreal scripts) to evaluate physical interaction capabilities \citep{hu2024scenecraft, lu2025ll3m}.

\section{Methodology}
 
The MCP4IFC framework connects LLMs with IFC-based BIM workflows via the Model Context Protocol. The framework offers a structured interface that allows LLMs to query, create, and edit IFC elements in a standardized way. This section explains the overall system architecture and the toolset that allows LLMs to interact directly with IFC data.

\subsection{System Architecture and Workflow}

The MCP4IFC framework consists of three main components, as shown in Figure \ref{overview_image}: the AI Client, the MCP Server, and the Blender add-on. The AI Client is an LLM interface that interacts with the user and displays the output. The MCP Server is a Python program that implements the MCP standard and manages all communication between the AI Client and the Blender add-on. It exposes a structured list of BIM tools, each defined as a JSON object that specifies its name, description, and input/output schemas. These tool definitions act as high-level wrappers, while the actual BIM operations are executed inside the Blender add-on. The add-on uses the IfcOpenShell library to manipulate IFC data and the Bonsai extension to update geometry in Blender’s 3D scene. Two communication links connect these components: the AI Client sends JSON-RPC messages to the MCP Server, and the server forwards commands to the Blender add-on through a socket connection.

\begin{figure}[tb]
  \centering
  \includegraphics[width=\linewidth]{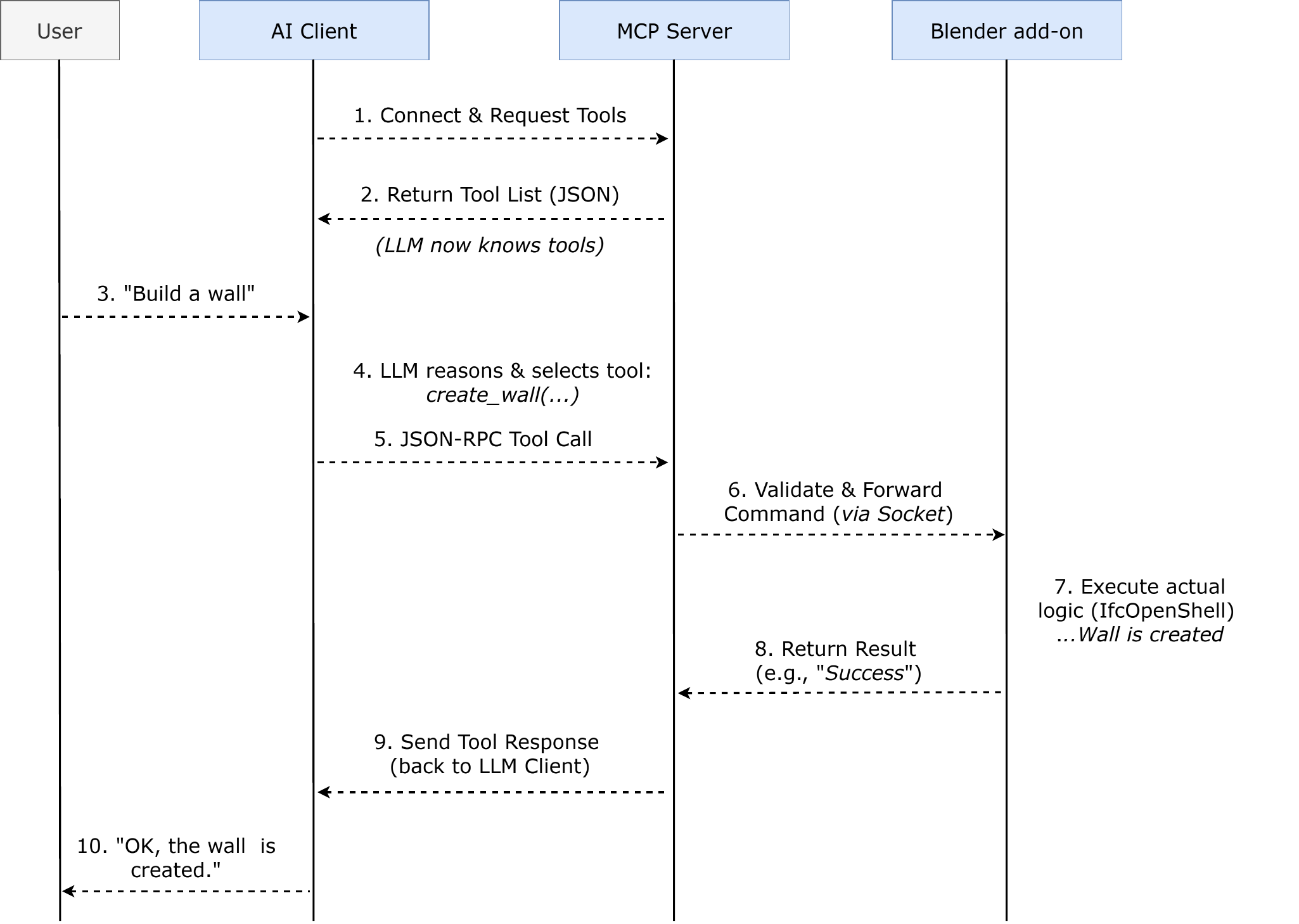}
  \caption{Workflow of a simple wall creation request through the MCP4IFC framework.}
  \label{overview_detailed}
\end{figure}

A detailed workflow is shown in Figure \ref{overview_detailed}. When the AI Client connects to the MCP Server, it requests the available tools. After receiving the tool list in JSON format, the client adds it to the system prompt, so the model can understand the available functions through in-context learning. When a user provides a prompt, such as \emph{add a 5-meter wall}, the LLM selects an appropriate tool, generates structured parameters, and sends a JSON-RPC request to the MCP server. The server validates this request and forwards it via the socket connection to the Blender add-on. The add-on executes the action and returns a structured result to the server, such as a success message or an error. The LLM analyzes it, determines the next step, and continues the loop until the task is completed.

\subsection{Toolset Implementation}

The MCP server includes a comprehensive list of tools. Each tool is defined with a clear name, description, input parameters, and expected outputs. The following sections describe the main tool categories and how they are implemented within the MCP4IFC framework.

\subsubsection{Scene Query and Context Tools}

Before an LLM can modify a BIM model, it needs to understand the existing elements in the scene. MCP4IFC provides scene and data querying tools for this purpose. These tools return a compact, structured summary of the model instead of exposing the entire IFC file. The summary lists each top-level element, including important fields such as the element’s GUID, IFC type, and basic properties, such as name, type, and location. This gives the LLM a high-level view of the scene that is easy to parse and reason about. Using this information, the model can answer questions like \emph{How many walls exist and what are their lengths?} or select the correct object to change based on a given identifier. This design offers several advantages over directly exposing IFC or related formats such as ifcXML or ifcJSON. Raw IFC data is long, highly detailed, and not efficient for LLMs to process. In contrast, the structured JSON summary retains only the information needed for LLM reasoning, laying the groundwork for future extensions that explore even more efficient LLM-oriented scene representations.


In addition to structured data, MCP4IFC supports multimodal inputs to provide spatial context. These tools capture screenshots of the modeling environment, either the full application window or only the 3D view, to show the model's current state. This helps the model verify changes visually, similar to how human designers work by observing the scene and refining results iteratively.


\subsubsection{Predefined IFC Element Creation tools}
\label{predefined-tools}

\begin{figure}[t!]
\centering
\captionsetup[subfigure]{justification=centering,font=footnotesize,labelformat=parens}

\begin{subfigure}{0.32\linewidth}
  \centering
  \includegraphics[width=\linewidth]{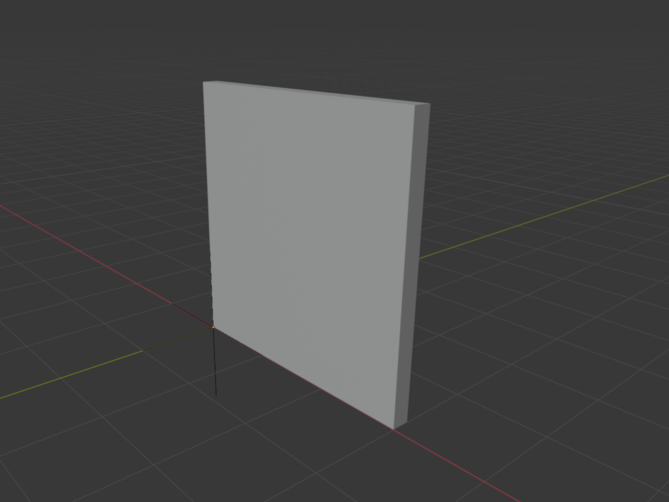}
  \caption{\emph{create\_wall()}}
\end{subfigure}\hfill
\begin{subfigure}{0.32\linewidth}
  \centering
  \includegraphics[width=\linewidth]{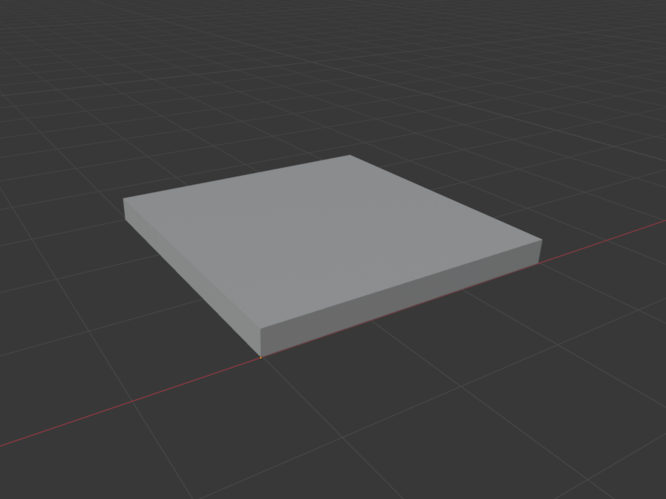}
  \caption{\emph{create\_slab()}}
\end{subfigure}\hfill
\begin{subfigure}{0.32\linewidth}
  \centering
  \includegraphics[width=\linewidth]{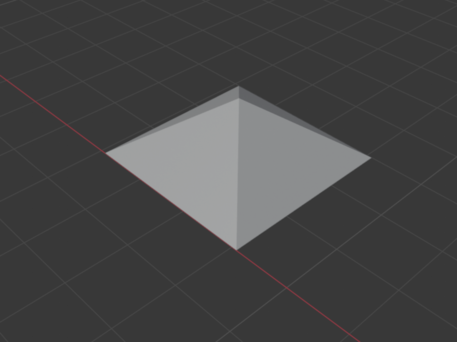}
  \caption{\emph{create\_roof()}}
\end{subfigure}

\vspace{6pt}

\begin{subfigure}{0.32\linewidth}
  \centering
  \includegraphics[width=\linewidth]{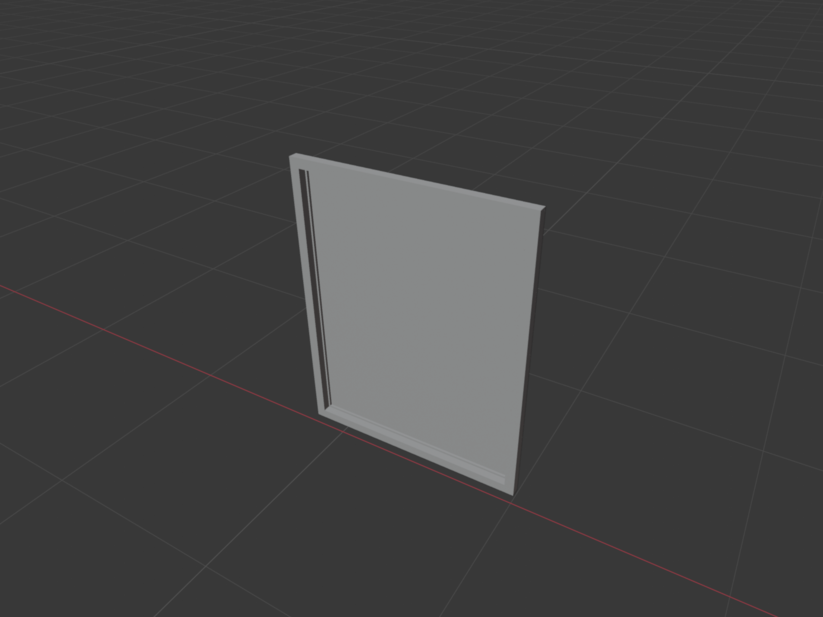}
  \caption{\emph{create\_window()}}
\end{subfigure}\hfill
\begin{subfigure}{0.32\linewidth}
  \centering
  \includegraphics[width=\linewidth]{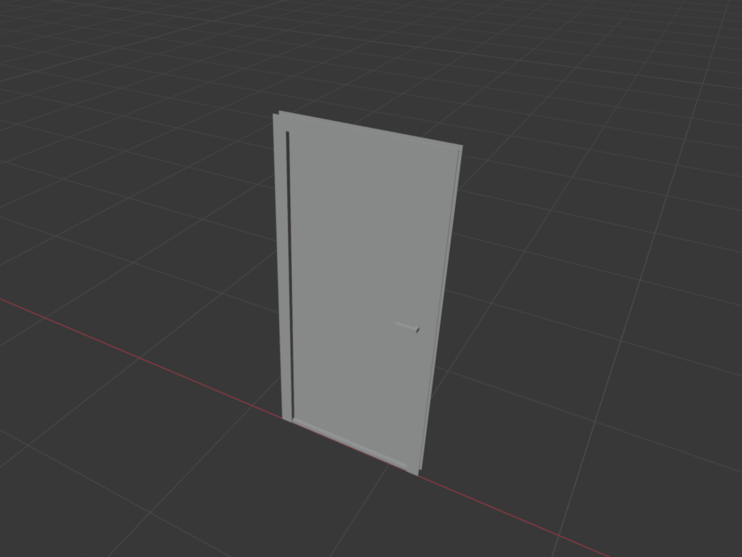}
  \caption{\emph{create\_door()}}
\end{subfigure}\hfill
\begin{subfigure}{0.32\linewidth}
  \centering
  \includegraphics[width=\linewidth]{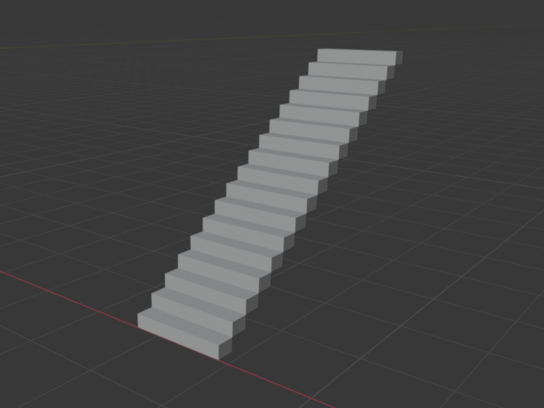}
  \caption{\emph{create\_stairs()}}
\end{subfigure}

\caption{ Example IFC building elements generated through predefined tool functions. These functions also support a variety of parametric input arguments.}
\label{fig:predefined_tool_examples}
\end{figure}

The LLM can use predefined tools to create, modify, or delete elements in the IFC model. These tools perform standard IFC operations on core building components, including walls, slabs, roofs, doors, windows, and stairs, as demonstrated in Figure \ref{fig:predefined_tool_examples}. Each tool has a set of well-defined input parameters that specify the properties and placement of the element. For example, the wall creation tool accepts a start and end point along with height and thickness values to generate an IFC wall element. Another variant of this tool takes a list of points and makes a chain of walls to outline a room. Slabs are created from closed polygon outlines and a thickness parameter, while roofs can be generated from a 2D outline or by referencing supporting walls. Doors and windows are inserted into existing walls by providing the wall GUID along with size and position. Every newly created element will have a unique IFC GUID, which is returned to the AI Client for continued reference in the next operations. In addition, the framework provides editing and deletion tools that enable an iterative workflow in which the model builds, checks, and refines the design step by step. The quality of the final design depends on how effectively the LLM selects the appropriate tools and configures their parameters. In practice, the LLM usually begins by creating a basic layout, then adds and refines elements using the predefined tools until the desired configuration is achieved.

\subsection{Dynamic Code generation tools}
\label{dynamic-code-generation-tools}

Predefined tools cover most of the common BIM operations, but they cannot represent every possible task. Real building models often require custom shapes, specific IFC manipulations, or unique design logic that goes beyond the fixed toolset. To support these cases, MCP4IFC includes dynamic code generation tools, which allow the LLM to write and execute Python scripts during inference. Each script runs inside a secure Blender environment and returns structured feedback to the AI Client. This enables handling new element types, advanced geometry generation, and specialized workflows without requiring manual development of new predefined tools. 

Dynamic tool usage falls into two main categories: IFC code generation and mesh generation. For operations involving IFC data that are not covered by existing tools, the LLM can directly generate Python code using the IfcOpenShell API. Since these libraries are publicly available online, LLMs have typically seen them during pretraining. This enables them to generate valid Python scripts that can manipulate IFC data directly. The code is executed in a sandboxed environment that blocks system access and restricts modules to approved libraries. After execution, the system returns logs, error messages, and object identifiers, which the LLM can analyze to refine and verify its operations. To further improve reliability, the MCP Server integrates a retrieval-augmented generation (RAG) system. It stores IfcOpenShell documentation, IFC schema definitions, and example scripts in a local vector store database. When needed, the LLM can retrieve relevant context using semantic search tools. This ensures the generated code uses the correct and up-to-date APIs.

Dynamic code generation also extends to direct geometry creation. Predefined IFC creation tools cannot easily represent complex elements such as curved ramps, bridges, or freeform structures. In such cases, the framework allows the LLM to generate Python code using external 3D geometry libraries. In principle, any compatible Python library can be used, as long as the LLM has sufficient knowledge of it from pretraining to generate correct API calls. The framework uses trimesh \citep{trimesh} for its simplicity, but other libraries can also be integrated \citep{khan2024text2cad}. The LLM determines the appropriate IFC class and a few semantic properties for the generated mesh, though it may not fully generate the IFC Space or the relationships between objects and this could be future work. This enables open-ended design where the LLM can create complex geometry procedurally. Future versions of the framework can incorporate more advanced parametric modeling libraries to further enhance complex and advanced design. 


\section{Experiments and Results}
The following experiments aim to demonstrate the feasibility and practical functioning of the MCP4IFC framework. 
Rather than providing an exhaustive quantitative benchmark, our goal is to illustrate that the proposed system can perform key BIM operations, i.e., querying, editing, and generation, through standardized tool use and dynamic code generation. The experiments serve as a proof of concept to validate the end-to-end workflow and to highlight potential research directions.

\subsection{Scene Querying}
\begin{figure}[htbp]
\centering
\begin{minipage}[c]{0.35\textwidth} 
    \centering
    \includegraphics[width=\linewidth]{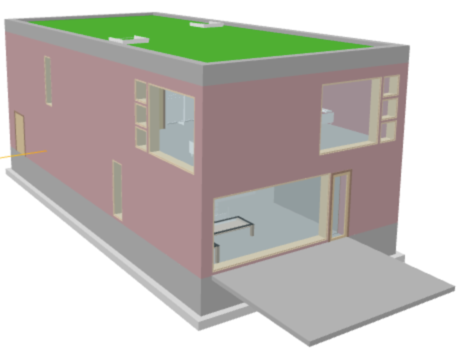}
    \caption{Duplex House}
    \label{fig:duplex}
\end{minipage}\hfill
\begin{minipage}[c]{0.60\textwidth} 
    \small
    \renewcommand{\arraystretch}{1.1}
    \setlength{\tabcolsep}{3pt} 
    {
    \rowcolors{2}{gray!5}{white}
    \begin{tabular}{|p{7.6cm}|} 
    \hline
    \textbf{Q:} Which rooms have direct access to outdoor spaces? \\ 
    \textbf{A:} [...] 4 Rooms have direct access to outdoor spaces: A101, B101, A102, B102. [...] 
    \\ \hline

    \textbf{Q:} Are the walls of room A203 load bearing? \\ 
    \textbf{A:} [...] The interior partition and plumbing walls in room A203 are NOT load-bearing [...]
    \\ \hline

    \textbf{Q:} What is the depth of the foundation? \\ 
    \textbf{A:} [...] The depth of the foundation is 0.3 meters (300 mm). [...]
    \\ \hline

    \textbf{Q:} Are all internal doors compliant with accessibility requirements? \\ 
    \textbf{A:} Based on my analysis of all 14 doors in your IFC project, not all internal doors are compliant with accessibility requirements. [...]
    \\ \hline
    \end{tabular}
    }
    \caption{Questions and correctly given IFC-based answers by Sonnet 4.5 on the ARC Duplex  using MCP4IFC.}
    \label{fig:ifc-answers}
\end{minipage}
\end{figure}

This section evaluates the performance of the MCP4IFC system on scene querying tasks.
For this, we use a subset of the IFC-bench-v1 dataset \citep{Hellin2025Natural} of $65$ questions associated with the \textit{Duplex} project's architectural (ARC) and mechanical, electrical and plumbing (MEP) models.
A visualization of the Duplex house can be seen in Figure \ref{fig:duplex}.
Following \cite{Hellin2025Natural}, the dataset can be divided into three categories of questions:
\begin{itemize}
  \item \textbf{Direct (27 tasks):} Queries solvable by directly extracting information from the IFC model (e.g., retrieving a specific object's GUID).
  \item \textbf{Indirect (16 tasks):} Queries requiring aggregation or computation (e.g., counting all radiators or calculating a total area).
  \item \textbf{Insufficient (22 tasks):} Queries where the necessary information is not present in the IFC model, potentially requiring the LLM to make assumptions or use its own expertise to solve the question (e.g., estimating how much of the materials can be recycled at the end-of-life).
\end{itemize}

We conduct the experiment on two LLMs, Claude Sonnet 4.5 and GPT-5 mini, and manually check answers for correctness, i.e. without using the LLM-as-Judge evaluation framework as conducted by \cite{Hellin2025Natural}.
As shown in Table \ref{tab:sq_results}, GPT-5 mini works best achieving an accuracy of $83.1\%$, compared to $75.4\%$ accuracy of Sonnet 4.5.
Hereby the benefit of the gpt model primarily stems from a better performance on \textit{direct} retrieval questions, solving $26$ tasks correctly compared to Claude's $21$.
In general, both models peform well on tasks with directly available information ($92.6\%/77.8\%$) and in reasoning on tasks for which the model contains insufficient information ($100\%/95.5\%$), showing that determining if information is stored inside the IFC model and reading it works well for current LLMs.
As expected, tasks with indirectly available information are most difficult as they require the model to retrieve and aggregate specific results.

We further compare the performance against the results reported by \cite{Hellin2025Natural} on Sonnet 3.5.
Note that the results are not directly comparable, because our system was only evaluated on a subset of IFC-Bench-v1, whereas \cite{Hellin2025Natural} evaluate on the whole dataset.
Although the functions of our MCP4IFC system are not adapted on the IFC-bench-v1 dataset tasks, the performance is similar to their system, which achieved an accuracy of $80\%$.

The tools mostly used by Sonnet 4.5 on correctly solved tasks can be seen in Table \ref{tab:sq_tool_distribution}.
Notably, the \emph{execute\_ifc\_code\_tool} function is called most often, which allows the model to write its own code.
This is often necessary to aggregate information, e.g. when requiring spatial relationships, counting or selecting objects of a distinct type.
Furthermore, this presents an important benefit over the tools defined by \cite{Hellin2025Natural}, as it allows the system to create its own task-specific tools to answer a query.

\begin{table}[h!]
\centering
\small
{
\rowcolors{2}{gray!5}{white}
\begin{tabular}{
    l
    c c r
    c c r
    r
}
\toprule
\multicolumn{1}{c}{\textbf{Category}} &
\multicolumn{3}{c}{\textbf{GPT-5 mini}} &
\multicolumn{3}{c}{\textbf{Sonnet 4.5}} &
\multicolumn{1}{c}{\textbf{Sonnet 3.5*}} \\
\cmidrule(lr){2-4} \cmidrule(lr){5-7} \cmidrule(lr){8-8}
 & \textbf{Correct} & \textbf{Incorrect} & \textbf{Accuracy}
 & \textbf{Correct} & \textbf{Incorrect} & \textbf{Accuracy}
 & \textbf{Accuracy} \\
\midrule
Direct        & 25 & 2 & 92.6\%
              & 21 & 6 & 77.8\%
              & 95\% \\
Indirect      & 7  & 9 & 43.8\%
              & 7  & 9 & 43.8\%
              & 62\% \\
Insufficient  & 22 & 0 & 100.0\%
              & 21 & 1 & 95.5\%
              & 73\% \\
\midrule
\textbf{Total} & \textbf{54} & \textbf{11} & \textbf{83.1\%}
               & \textbf{49} & \textbf{16} & \textbf{75.4\%}
               & \textbf{80\%} \\
\bottomrule
\end{tabular}
}
\caption{Comparison of scene querying results between \textbf{GPT-5 mini}, \textbf{Sonnet 4.5}, and \textbf{Sonnet 3.5}. 
*Results for Sonnet 3.5 are reported by \cite{Hellin2025Natural}, evaluated on a larger dataset and therefore not directly comparable.}
\label{tab:sq_results}
\end{table}

\begin{table}
\centering
{
\rowcolors{2}{gray!5}{white}
\begin{tabular}{lr}
\toprule
tool & count \\
\midrule
execute\_ifc\_code\_tool & 91 \\
get\_ifc\_scene\_overview & 45 \\
get\_scene\_info & 29 \\
get\_object\_info & 16 \\
execute\_blender\_code & 7 \\
search\_ifc\_knowledge & 2 \\
get\_door\_properties & 1 \\
capture\_blender\_3dviewport\_screenshot & 1 \\
\bottomrule
\end{tabular}
}
\caption{Distribution of tool usage among the correct solutions.}
\label{tab:sq_tool_distribution}
\end{table}

Our evaluation comes with important limitations. First, the binary correct/incorrect metric can be insufficient for certain ambiguous questions where an answer might be partially correct. In our evaluation, we optimistically marked such cases as correct, suggesting our reported accuracies represents an upper bound. A more nuanced evaluation scheme may be necessary for future work. 
Second, grouping tasks into the proposed categories by \cite{Hellin2025Natural} led to difficulties, as, e.g., tasks with indirectly available information often contain to a high proportion of subtasks where direct retrieval is required, and vice versa, making a clean categorization difficult.
However, we grouped tasks to the best of our knowledge and, given similar performance to \cite{Hellin2025Natural}, think that our assessment is close to the unreleased ground truth.
Lastly, as the MCP4IFC system offers a variety of different tools to query and generate BIM structures, the system currently requires a large context size in the input prompt.
As this directly represents a higher cost and less than $10$ of the tools were actually required to successfully solve tasks in scene querying (see Table \ref{tab:sq_tool_distribution}), it is advisable to generally disable generative tools for such a task and vice versa.

\subsection{Editing and Generation Experiments}

In addition to the querying benchmark, we evaluated how MCP4IFC performs on editing and generation tasks. These experiments test whether the system can follow high-level instructions and perform modeling operations without manual GUI interaction. The experiments cover two main tasks: (1) semantic IFC editing via code generation, where the LLM writes and executes Python code to modify IFC data, and (2) IFC geometry generation, where natural language prompts are translated into full 3D building elements. All experiments were conducted using the Claude Sonnet 4.5 model in the Claude Desktop interface. 


\subsubsection{IFC Editing via Code Generation (Semantic/Non-Geometric)}
\label{IFC-editing-code-gen-exp}

When predefined tools cannot complete a request, the LLM generates new Python code to perform the task (as described in Section \ref{dynamic-code-generation-tools}). This experiment focuses on semantic and non-geometric IFC modifications, for which there are no specialized tools. We designed eight editing tasks (Table \ref{tab:edit_codegen}) that included batch renaming, creating property sets, classifying elements, and performing custom calculations. We prompted the LLM to perform these operations using our MCP framework and analyzed whether it could complete the task.


All experiments were performed on the \emph{BasicHouse.ifc} model from the BIM-Whale repository \citep{andrewisen2025bimwhaleifcsamples}. Claude Sonnet 4.5 completed all eight editing tasks via dynamic code generation. Each result was manually verified by comparing the IFC file before and after execution. Although the task set is limited, the results show that the LLM can reliably execute semantic edits and generate valid IFC-compatible code. More complex cases may fail when geometry, spatial context, or element dependencies are required. Overall, these experiments demonstrate that direct IFC editing via LLM-driven code generation is feasible for a range of semantic operations.


\begin{table}[tb]
\centering
\small
\setlength{\tabcolsep}{4pt}        
\renewcommand{\arraystretch}{1.15}
\caption{The eight prompts used for the IFC editing experiments via dynamic code generation. Claude Sonnet 4.5 completed all tasks using the MCP server. The RAG column indicates whether the LLM invoked the retrieval tool.}
\label{tab:edit_codegen}
{
\rowcolors{2}{gray!5}{white}
\begin{tabularx}{\linewidth}{@{}>{\raggedright\arraybackslash}X
                                >{\centering\arraybackslash}m{0.20\linewidth}
                                >{\centering\arraybackslash}m{0.10\linewidth}
                                >{\centering\arraybackslash}m{0.10\linewidth}@{}}
\toprule
\textbf{Prompt} & \textbf{Input IFC} & \textbf{RAG} & \textbf{Outcome} \\
\midrule
1.\ Rename all selected walls to include their height in the name (e.g., ``Wall-3.0m''). & \texttt{BasicHouse.ifc} & No & Success \\
2.\ Change the description of the building to ``High-rise residential tower''. & \texttt{BasicHouse.ifc} & No & Success \\
3.\ Update the name of all \texttt{IfcDoor} objects by appending their floor level to the existing name. & \texttt{BasicHouse.ifc} & No & Success \\
4.\ Add a custom property set ``Thermal\_Properties'' to all walls with properties: U-value = 0.25, Insulation\_Type = ``Mineral Wool''. & \texttt{BasicHouse.ifc} & Yes & Success \\
5.\ Find all slabs and add a ``Fire\_Rating'' property with value ``2HR''. & \texttt{BasicHouse.ifc} & Yes & Success \\
6.\ Classify all walls and slabs in the IFC model using Uniclass 2015 (walls: \texttt{Ss\_25\_10\_20}, slabs: \texttt{Ss\_25\_30}). & \texttt{BasicHouse.ifc} & Yes & Success \\
7.\ Add cost information to all windows (unit cost \$500) and compute the total based on the window count. & \texttt{BasicHouse.ifc} & No & Success \\
8.\ Update the owner history for selected elements with current timestamp and user ``BIM Manager''. & \texttt{BasicHouse.ifc} & Yes & Success \\
\bottomrule
\end{tabularx}
}
\end{table}

\subsubsection{IFC Geometry Generation}

\begin{figure}[t!]
\centering
\captionsetup[subfigure]{justification=centering,font=footnotesize,labelformat=parens}

\begin{subfigure}{0.32\linewidth}
  \centering
  \includegraphics[width=\linewidth]{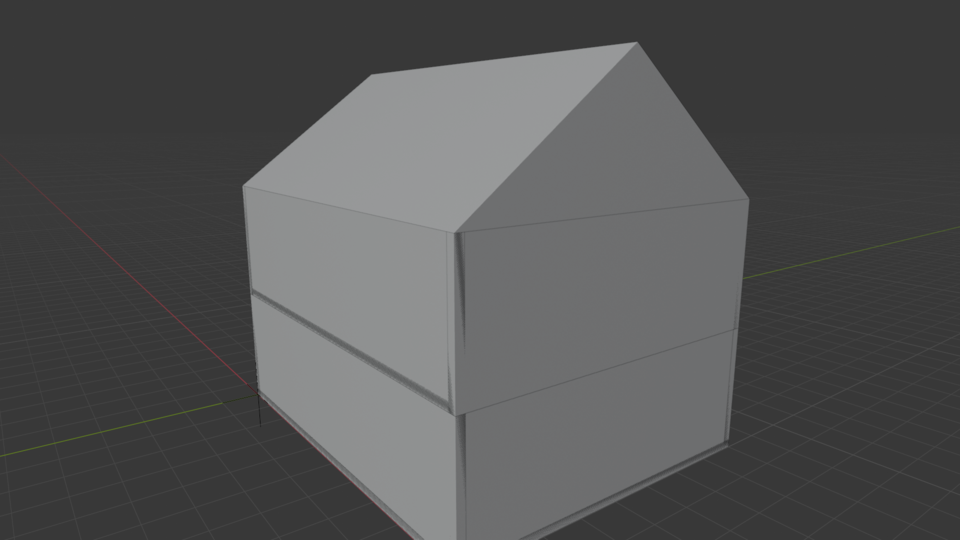}
  \caption{Two-storey structure}
\end{subfigure}\hfill
\begin{subfigure}{0.32\linewidth}
  \centering
  \includegraphics[width=\linewidth]{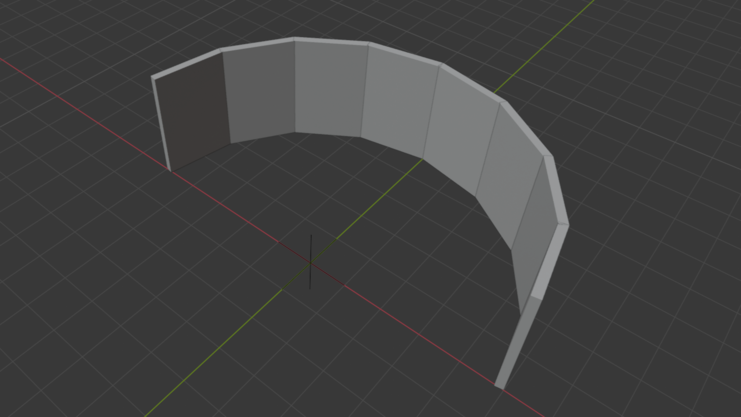}
  \caption{Curved wall (8 segments)}
\end{subfigure}\hfill
\begin{subfigure}{0.32\linewidth}
  \centering
  \includegraphics[width=\linewidth]{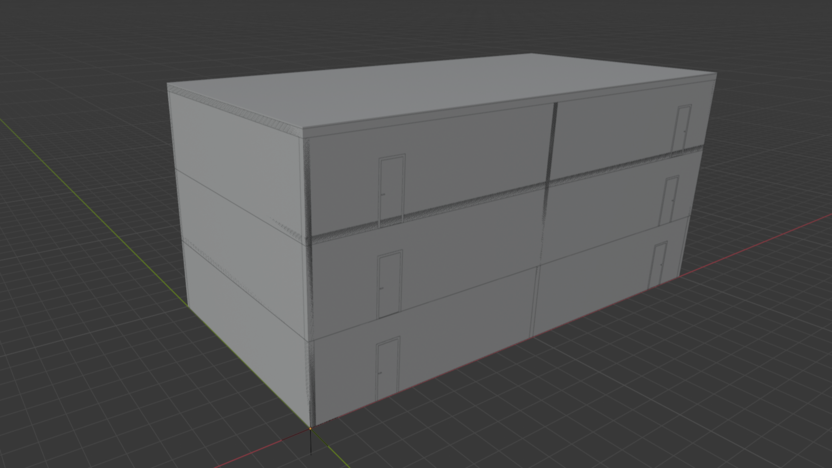}
  \caption{3-storey apartment block}
\end{subfigure}

\vspace{6pt}

\begin{subfigure}{0.32\linewidth}
  \centering
  \includegraphics[width=\linewidth]{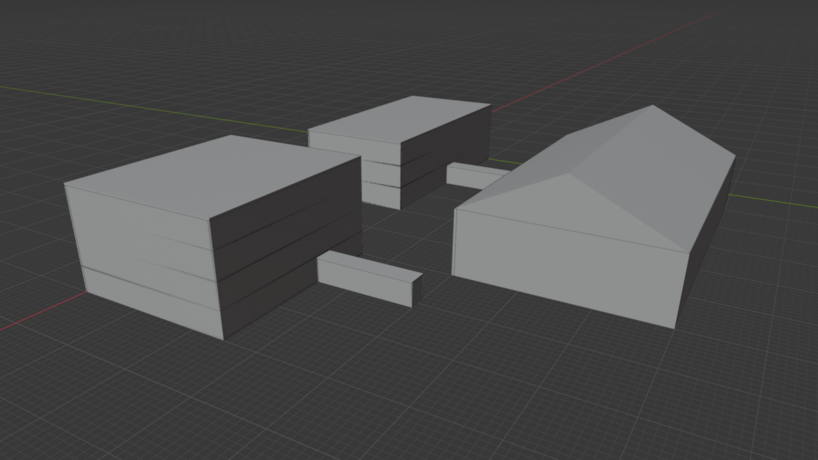}
  \caption{University campus (3 blocks)}
\end{subfigure}\hfill
\begin{subfigure}{0.32\linewidth}
  \centering
  \includegraphics[width=\linewidth]{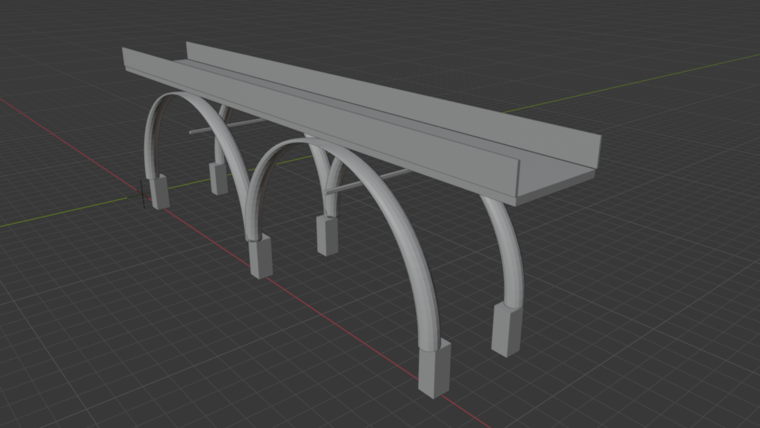}
  \caption{Landmark bridge}
\end{subfigure}\hfill
\begin{subfigure}{0.32\linewidth}
  \centering
  \includegraphics[width=\linewidth]{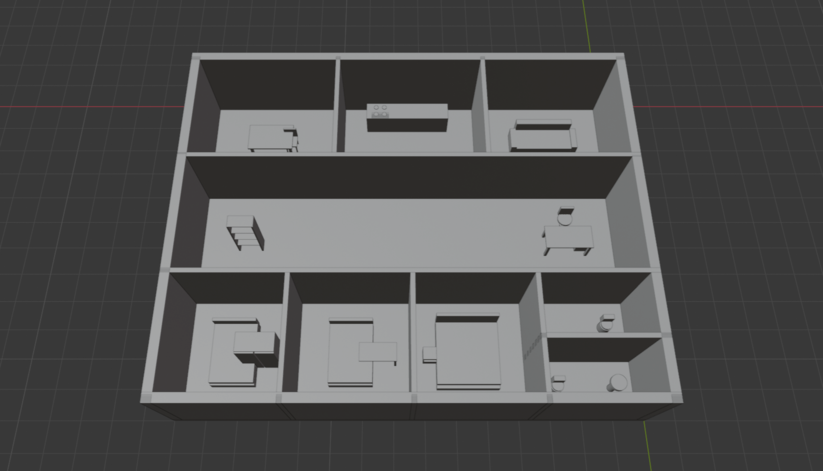}
  \caption{10-room single-floor house}
\end{subfigure}

\caption{IFC object generation from textual prompts.
Examples (a-f) are generated using the tools available in the MCP server. Here are the detailed prompts: 
\textit{(a)} Build a two-story structure: ground floor slab 10 m × 8 m, four perimeter walls 3 m high, second floor slab at 3 m, another set of walls 3 m high, and a gable roof. 
\textit{(b)} Create a curved wall approximation with 8 segments forming a semi-circle, radius 5 m, height 3 m. 
\textit{(c)} Create a three-storey residential apartment block with two flats per floor, each flat having a living room, kitchen, bedroom, and bathroom. 
\textit{(d)} Create a university campus with three buildings: a main lecture block, a library, and a sports hall connected by covered walkways. 
\textit{(e)} Generate a landmark bridge structure with arch supports and pedestrian deck. 
\textit{(f)} Generate a single-floor house with ten rooms, each furnished with appropriate interior objects (e.g., bed, table, or sofa).}
\label{fig:singleprompt_design}
\end{figure}

\begin{figure}[t!]
\centering
\captionsetup[subfigure]{justification=centering,font=footnotesize,labelformat=parens}

\begin{subfigure}{0.32\linewidth}
  \centering
  \includegraphics[width=\linewidth]{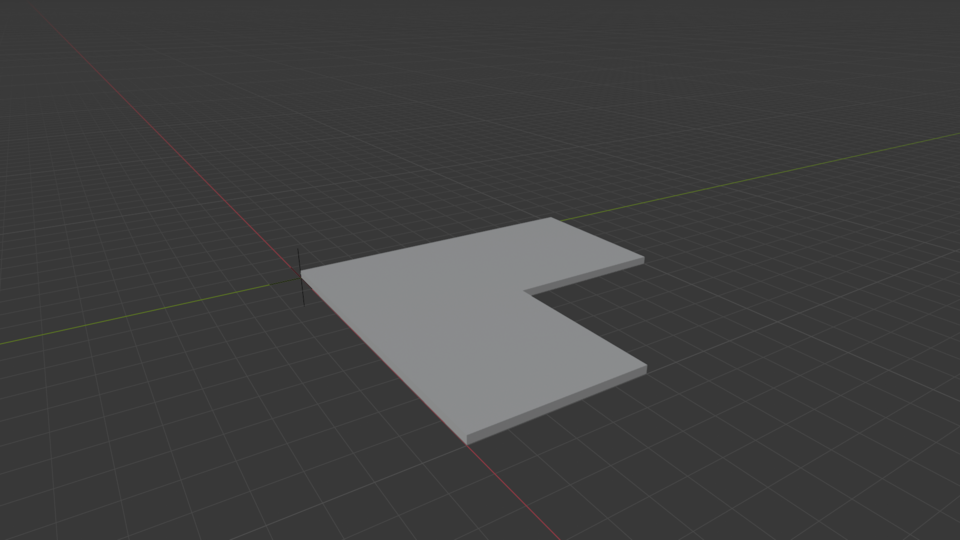}
  \caption{L-shaped slab foot}
\end{subfigure}\hfill
\begin{subfigure}{0.32\linewidth}
  \centering
  \includegraphics[width=\linewidth]{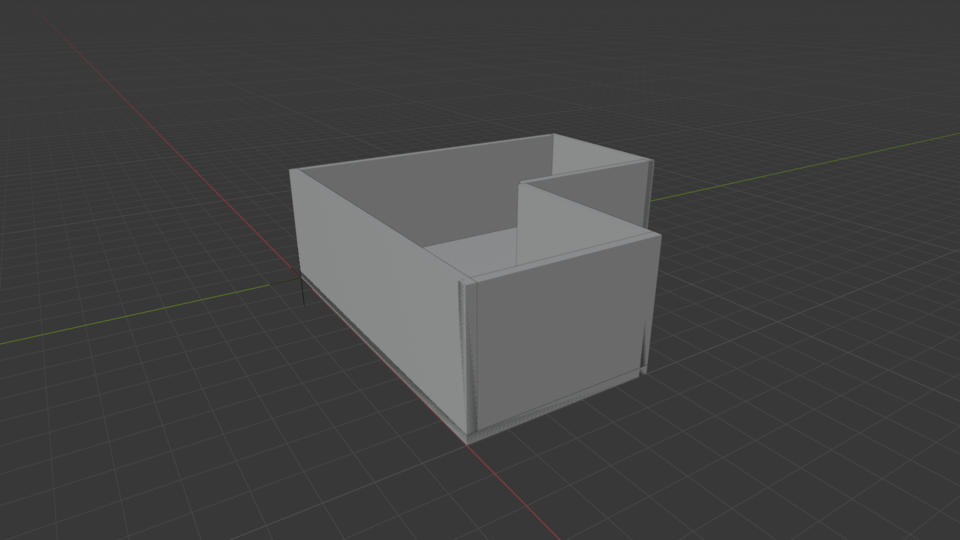}
  \caption{Perimeter walls added}
\end{subfigure}\hfill
\begin{subfigure}{0.32\linewidth}
  \centering
  \includegraphics[width=\linewidth]{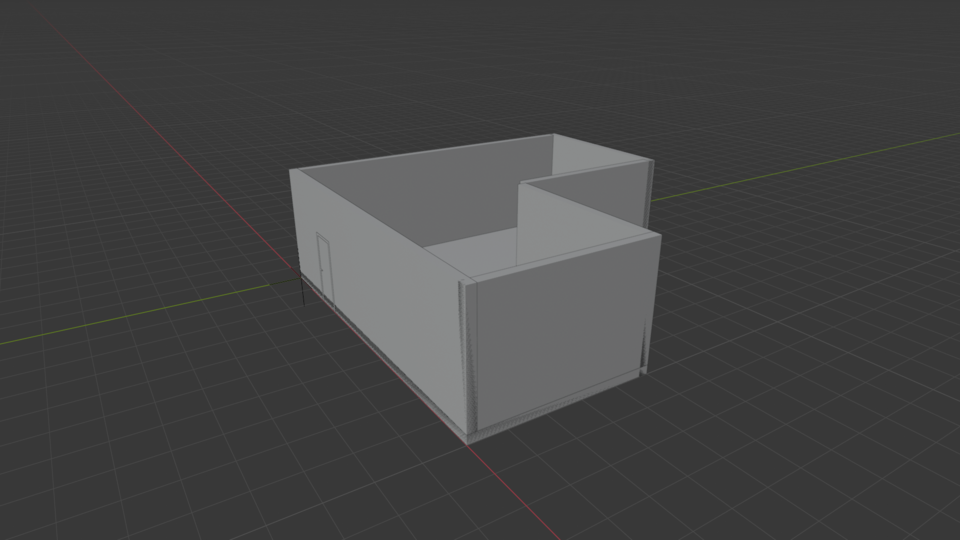}
  \caption{Door insertion}
\end{subfigure}

\vspace{6pt}

\begin{subfigure}{0.32\linewidth}
  \centering
  \includegraphics[width=\linewidth]{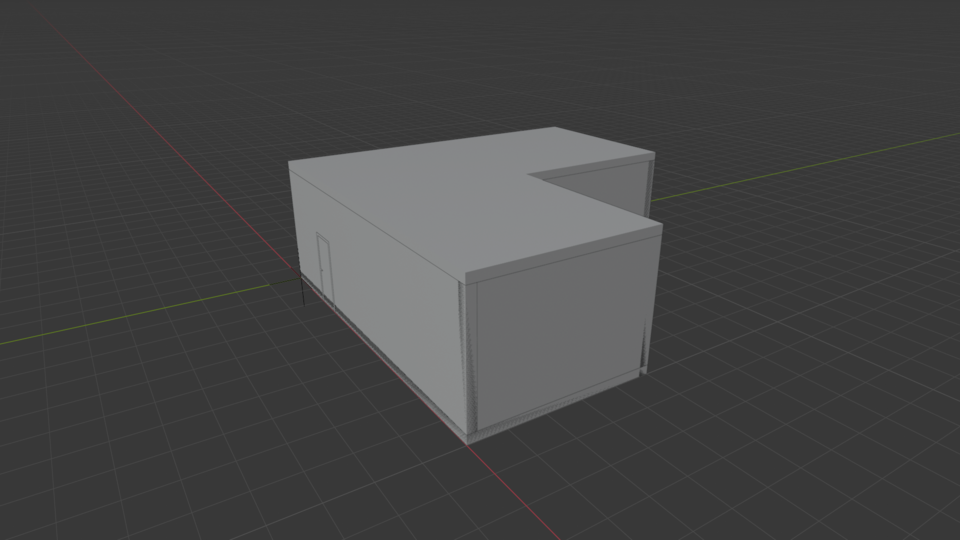}
  \caption{Second-floor slab}
\end{subfigure}\hfill
\begin{subfigure}{0.32\linewidth}
  \centering
  \includegraphics[width=\linewidth]{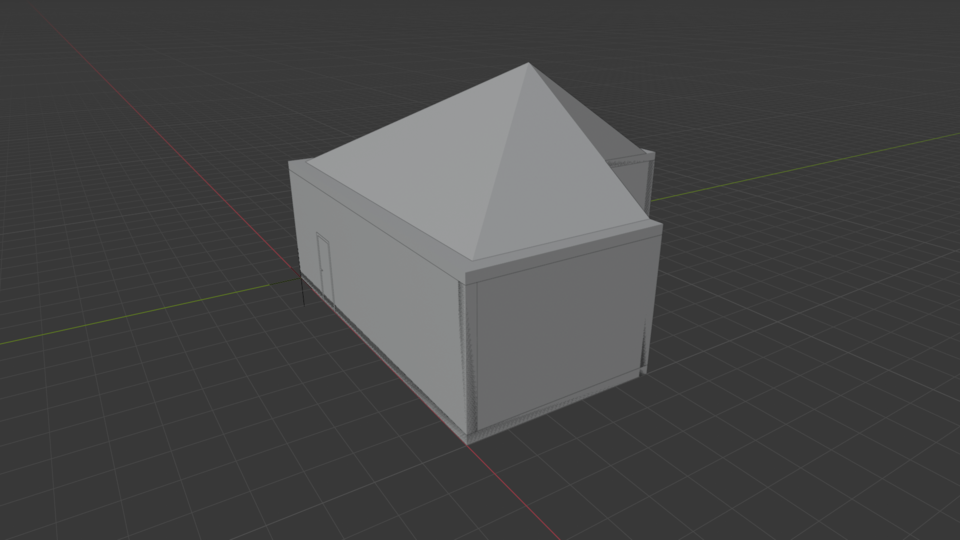}
  \caption{Hip roof generation}
\end{subfigure}\hfill
\begin{subfigure}{0.32\linewidth}
  \centering
  \includegraphics[width=\linewidth]{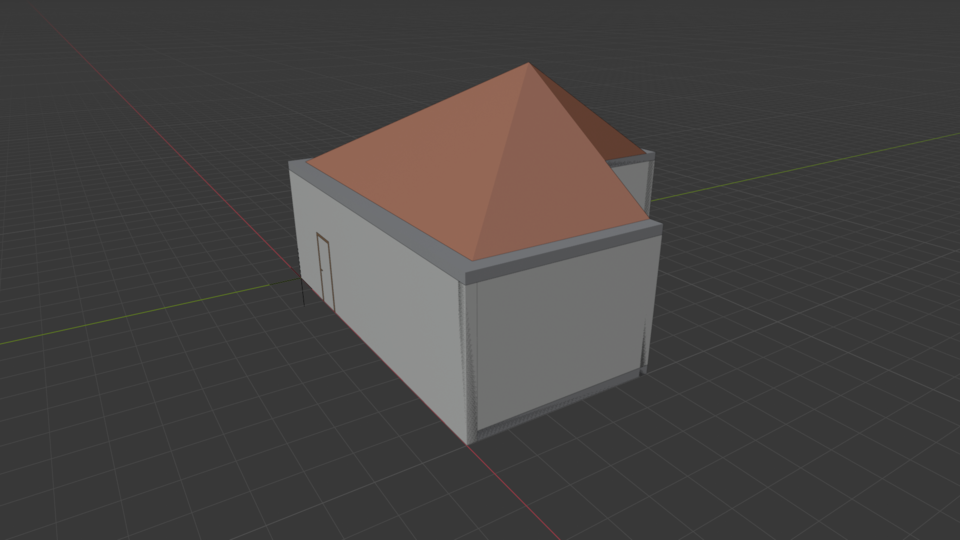}
  \caption{Styled final building}
\end{subfigure}

\caption{Step-by-step generation of an L-shaped building through sequential instructions. Here are the detailed prompts:
\textit{(a)} Create an L-shaped floor slab using polyline: vertices at (0,0), (10,0), (10,5), (5,5), (5,10), (0,10), thickness 0.25m at height 0.
\textit{(b)} Create walls along this L-shaped perimeter: height 3.5m, thickness 0.25m, following the slab outline. 
\textit{(c)} Add a door at position (2, 0, 0) - entrance on the south side.
\textit{(d)} Create another L-shaped slab at height 3.5m matching the ground floor shape.
\textit{(e)} Add a hip roof covering the entire L-shaped building with 30-degree slope.Generate a landmark bridge structure with arch supports and pedestrian deck. 
\textit{(f)} Can you style the house.Generate a single-floor house with ten rooms, each furnished with appropriate interior objects (e.g., bed, table, or sofa).}
\label{fig:multiturn_design}
\end{figure}

We evaluated the model’s ability to generate IFC geometry both from single prompts and through multi-turn interactive workflows. Geometry creation in MCP4IFC is supported by predefined high-level tools \ref{predefined-tools} and by dynamic code generation \ref{dynamic-code-generation-tools}. As shown in Figure~\ref{fig:singleprompt_design}, the system can create a variety of structures from textual descriptions, demonstrating that the model can interpret high-level design intent and translate it into valid IFC geometry. We further analyzed sequential design workflows in which the LLM iteratively refined a single IFC model across multiple user instructions. Figure \ref{fig:multiturn_design} shows an example where the model preserved spatial context over six consecutive prompts to produce a coherent layout. This indicates strong potential for conversational BIM workflows where users can iteratively shape a design through natural language instead of manual GUI operations.

However, a closer review of the generated IFC files reveals some issues. In the example \emph{(f)} shown in Figure~\ref{fig:multiturn_design}, the geometry appears structurally valid, but the underlying IFC semantics were not fully specified. Walls did not connect properly at intersections, openings for windows and doors were not properly added, and rooms lacked \texttt{IfcSpace} definitions. Elements often lacked property sets and material assignments, and furniture lacked spatial containment relationships. These issues show that current models can generate geometric elements but struggle to maintain the relational and semantic structure required for reliable BIM workflows.

Overall, the results demonstrate that LLM-driven IFC generation is feasible but not yet semantically robust. Improving spatial reasoning, semantic consistency, and dependency handling remains essential. Future work should focus on fixing incomplete IFC structures and training models with better exposure to well-structured IFC data so they can learn to create better BIM models.

\section{Conclusion \& Future Work}
In this work, we presented MCP4IFC, an MCP server for generating, editing, and querying IFC models built on top of the IfcOpenShell API. The framework connects LLMs to BIM workflows through standardized MCP tool calls, enabling direct interaction with IFC data in an interoperable and transparent way. By combining predefined tool interfaces with dynamic code generation and retrieval-augmented learning, the system allows LLMs to perform a wide range of modeling and information retrieval tasks without depending on proprietary software APIs. Our experiments demonstrate that the proposed framework is capable of performing realistic BIM operations, from IFC querying to semantic editing and procedural geometry generation. While these experiments are not intended as comprehensive benchmarks, they demonstrate that the MCP4IFC framework effectively enables LLMs to perform end-to-end IFC manipulation through a standardized, extensible tool interface. MCP4IFC is released as open source to support further research in AI-assisted BIM design. We believe it provides a solid foundation for developing and evaluating future systems that integrate language models, spatial reasoning, and open standards to advance automation in the built environment.
Concretely, various research directions can be derived from this:

\textbf{Expanded and Efficient Tool Library:}
The current toolset can be further extended to support higher-level modeling operations. 
New tools for common building components, such as staircases, elevators, and facades, would allow AI agents to complete complex design actions with a single tool call. 
In addition, community experts could contribute their own extensions, which would become immediately accessible to any connected AI client through the standardized MCP interface.
Beyond predefined modeling tools, generative design capabilities can also be integrated. 
For instance, a floor-plan understanding model \citep{Wei2025PointOL, Fan_2022_CVPR} could transform images or sketches into room layouts, while advanced mesh-generation algorithms \citep{Wang2024LLaMAMeshU3, Hao2024MeshtronHA} could be combined with an LLM to convert complex geometries into structured IFC elements.
Together, these additions would enable model-based design exploration within IFC-compliant workflows.
At the same time, improving efficiency is essential: mechanisms that dynamically preselect which tools are exposed to the LLM, depending on the current task, could significantly reduce computational overhead (e.g., information retrieval tasks may not require generative modeling tools). Further research should also investigate more compact prompting strategies, such as minimizing redundant tool descriptions or providing only the most relevant tokens. Currently, the full toolset definitions already occupy around 40k tokens, and the context grows further for complex models, underscoring the need for lightweight and adaptive prompt representations.
Together, these additions would enable model-based design exploration within IFC-compliant workflows.

\textbf{Agentic BIM Workflows and AI Model Improvement}
The current framework supports individual modeling actions, but full automation requires higher-level planning and decision-making. Future research could develop BIM-specific agents that can plan complete tasks, monitor progress, and react to feedback. Improving the underlying LLMs is also essential. A model fine-tuned on BIM data, tool-call traces, and code execution logs could reason more effectively about geometry and semantics. Such a domain-specialized model could run locally to reduce cost and latency. Together, these advances would allow the agent to autonomously design, check, and refine IFC models through long-horizon reasoning.
In addition, current LLMs and VLMs still face fundamental limitations in geometric and topological understanding \citep{yin2025spatialmentalmodelinglimited, mao2025spatiallmtraininglargelanguage}. Future research could explore multi-modal spatial feedback mechanisms, allowing models to reason about geometry through structured representations such as graphs, point clouds, and visual embeddings \citep{perception_encoder, VJEPA2, DINOWM}. Integrating these world-model concepts from robotics could enhance spatial awareness and enable more reliable generative design, collision detection, and geometric reasoning in BIM workflows.

\textbf{Extending to other BIM software:} 
Most tools in the current framework are implemented through IfcOpenShell, with Bonsai and Blender serving as lightweight interfaces that provide a 3D modeling workspace and interactive editing capabilities. 
Because the core functionality is built on IfcOpenShell, these tools can be easily adapted to any BIM software environment that supports IFC import, modification, or visualization. 
This design makes the framework highly flexible and straightforward to integrate into diverse BIM platforms.

\section{Acknowledgements}
We would like to thank Juan David Rodriguez for valuable discussions and support regarding the MCP server integration, and Sylvain Hellin for providing the IFC-Bench-v1 dataset and for the insightful exchange. 
We also acknowledge the IfcOpenShell community for developing and maintaining the open platform on which this work builds. This research was supported by the Federal Ministry of Research, Technology and Space via the Show2Instruct project (Grant No. 01MK25008D).



\newpage

\bibliographystyle{iclr2026_conference}

\bibliography{bibliography, links}

\newpage

\appendix
\section{Appendix}

\subsection{Dynamic code Generation example}
This section provides an example of a script generated by the LLM during the semantic (non-geometric) IFC editing experiments described in Section~\ref{IFC-editing-code-gen-exp}. When a required operation is not covered by predefined tools, the LLM writes Python code that is executed inside the secure Blender sandbox. Listing \ref{lst:orig_python} shows a script generated by Claude Sonnet~4.5 to update the description of the building entity. Before writing the code, the model may call scene-query tools. These tools provide additional context, such as the ID of selected objects and the scene overview. After it has enough information, the model generates the code and uses the \emph{execute\_ifc\_code\_tool()} to execute the generated script. The script first loads the current IFC file that is already open in the Bonsai Blender environment using the \emph{get\_ifc\_file()} function. Then it performs the operation and uses the \emph{save\_and\_load\_ifc()} function to write the changes back to the IFC file and refreshes the model. This shows how the system can edit IFC data through code without any GUI actions.

\begin{lstlisting}[style=pystyle, 
    caption={LLM-generated script to change the description of the building to 'High-rise residential tower' (executed in Blender sandbox).}, 
    label={lst:orig_python}]
import ifcopenshell
import ifcopenshell.api
from blender_addon.api.ifc_utils import get_ifc_file, save_and_load_ifc

# Get the IFC file
ifc_file = get_ifc_file()  
# Find the building by GUID
building_guid = "3UdjywU2L4v9tTcFvuqwGm"
building = ifc_file.by_guid(building_guid)
# Check current description
print(f"Building name: {building.Name}")
print(f"Current description: {building.Description}")
# Edit the building's description attribute
ifcopenshell.api.run(
    "attribute.edit_attributes",
    ifc_file,
    product=building,
    attributes={"Description": "High-rise residential tower"}
)
# Verify the change
print(f"New description: {building.Description}")
# Save the changes
save_and_load_ifc()
print("Building description successfully updated!")
\end{lstlisting}

\subsection{Scene Query output example}

To demonstrate the LLM's scene-understanding capability, we created a simple test scene in an IFC file containing four walls in a 10x10m grid. The LLM was then prompted to summarize this scene. Listing \ref{lst:json_summary} shows the LLM output and Listing \ref{lst:json_outout} shows the raw JSON data retrieved by the LLM's tool call. 

\begin{lstlisting}[style=mdstyle, float, caption={The LLM generated output for the prompt - can you give a summary of the existing scene, use the get\_scene\_info function. The model calls the tool (<< {tool\_call}>>) and then summarizes the tool's raw JSON output.}, label={lst:json_summary}]
I will get the current scene information for you.
<<tool_call: get_scene_info>>

Scene Summary
The current Blender scene contains a basic IFC project structure with 9 objects total:
IFC Spatial Structure
- IfcProject - My Project (root container)
- IfcSite - My Site
- IfcBuilding - My Building
- IfcBuildingStorey - My Storey (active floor level)
Building Elements
- 4 Walls forming a rectangular enclosure:
  - Wall_001 at origin (0, 0, 0)
  - Wall_002 at (10, 0, 0)
  - Wall_003 at (10, 10, 0)
  - Wall_004 at (0, 10, 0)
  These appear to create a 10m x 10m room layout.
Type Objects
- IfcWallType - wall (template/type definition, currently hidden)

All objects are currently visible except for the wall type definition. No objects are currently selected. The scene represents a simple rectangular room structure within a standard IFC hierarchical organization.
\end{lstlisting}

\begin{lstlisting}[style=pystyle, float, caption={Raw JSON output from the get\_scene\_info tool call. This data structure lists all scene objects, including their properties and unique GUIDs. Note that even in a simple 4-wall scene, the output includes hierarchical elements (e.g., IfcProject, IfcBuildingStorey) as separate objects, since they have their own GUIDs. The output is truncated here.}, label={lst:json_outout}]
{
  "count": 9,
  "total": 9,
  "offset": 0,
  "limit": 9,
  "objects": [
    {
      "name": "IfcProject/My Project",
      "type": "EMPTY",
      "location": [0.0,0.0,0.0],
      "visible": true,
      "selected": false,
      "guid": "0CsEa6q1jESwVHd0H5grPl",
      "ifc_class": "IfcProject"
    },
    {
      "name": "IfcBuildingStorey/My Storey",
      "type": "EMPTY",
      "location": [0.0,0.0,0.0],
      "visible": true,
      "selected": false,
      "guid": "3POxSr5YnFlhRhdFQxAINX",
      "ifc_class": "IfcBuildingStorey"
    },
    {
      "name": "IfcWall/Wall_001",
      "type": "MESH",
      "location": [0.0,0.0,0.0],
      "visible": true,
      "selected": false,
      "guid": "3Cqrj0WO5Erf3j77QECM8D",
      "ifc_class": "IfcWall"
    },
    {
      "name": "IfcWall/Wall_002",
      "type": "MESH",
      "location": [10.0,0.0,0.0],
      "visible": true,
      "selected": false,
      "guid": "23EPa7DQvF2foQfEeTNPD_",
      "ifc_class": "IfcWall"
    },
    ....
\end{lstlisting}

\subsection{Floor Plan Generation from images}

\begin{figure}[tb]
\centering
\begin{minipage}{0.92\linewidth}
\centering

\begin{subfigure}{0.30\linewidth}
  \centering
  \includegraphics[width=\linewidth]{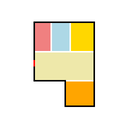}
\end{subfigure}\hfill
\begin{subfigure}{0.32\linewidth}
  \centering
  \includegraphics[width=\linewidth]{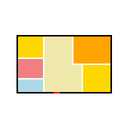}
\end{subfigure}\hfill
\begin{subfigure}{0.30\linewidth}
  \centering
  \includegraphics[width=\linewidth]{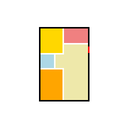}
\end{subfigure}

\vspace{8pt}

\begin{subfigure}{0.25\linewidth}
  \centering
  \includegraphics[width=\linewidth]{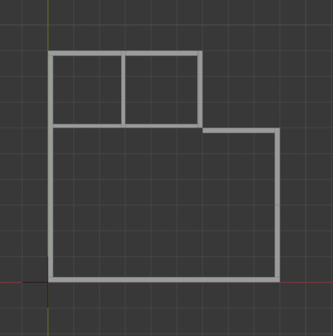}
\end{subfigure}\hfill
\begin{subfigure}{0.32\linewidth}
  \centering
  \raisebox{0.05\height}{\includegraphics[width=\linewidth]{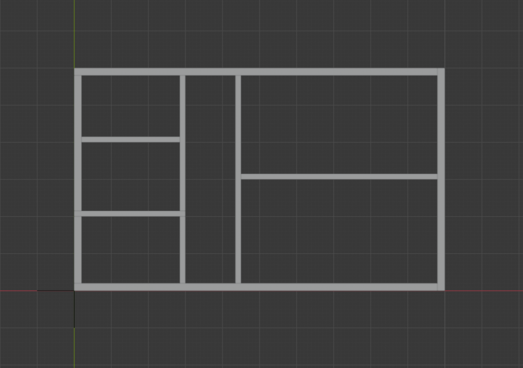}}
\end{subfigure}\hfill
\begin{subfigure}{0.25\linewidth}
  \centering
  \includegraphics[width=\linewidth]{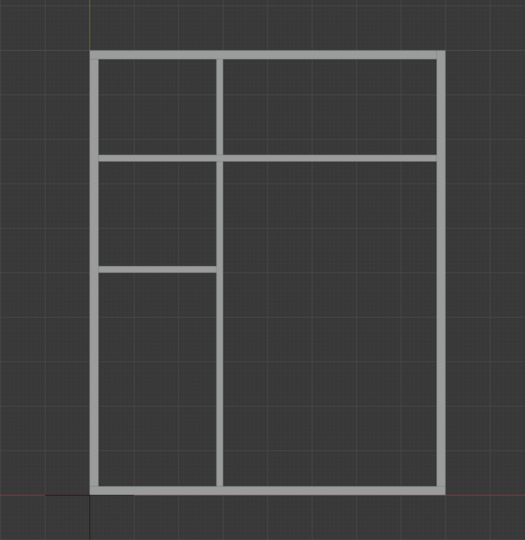}
\end{subfigure}

\end{minipage}

\caption{Floor plan generation using a multimodal LLM. The top row shows input floor plans from the Tell2Design dataset \cite{Leng2023Tell2DesignAD}, and the bottom row shows the corresponding wall layouts generated by the MCP server.}
\label{fig:floorplan_understanding}
\end{figure}

This experiment explores whether the multimodal abilities of LLMs could be used for floor plan understanding. Three floor-plan images from the Tell2Design dataset \citep{Leng2023Tell2DesignAD} were provided separately, and the LLM (Claude Sonnet 4.5) was prompted to generate the corresponding wall layouts. The input and output are shown in Figure \ref{fig:floorplan_understanding}. The LLM captured the overall structure but struggled with the spatial organization. Fine-grained geometry and more intricate layout features were often missing or incorrectly interpreted. These suggest that while multimodal LLMs can extract high-level spatial features from architectural drawings, they currently lack the precision needed for accurate reconstruction. Integrating a specialized floor-plan recognition model \citep{Fan_2022_CVPR, Wei2025PointOL} could improve performance by providing a more reliable extraction of boundaries and room structures.

\subsection{Visual Feedback Check using Screenshot Tools}

To explore whether a multimodal LLM can interpret BIM scenes directly from images, we ran a small experiment. The goal of this experiment was to assess whether the model could understand the 3D scene shown in each image and support basic visual reasoning tasks. We captured screenshots from different IFC models and asked the model a set of questions about visible elements. These questions can be categorized into three common types:

\begin{itemize}
    \item \textbf{Counting (25 questions):} Questions regarding the number of elements, such as walls, chairs, windows, or other elements, found in BIM.
    \item \textbf{Spatial Relationships (9 questions):} Questions about the relative positioning of objects, e.g., "Where is the TV located relative to the sofa?"
    \item \textbf{Attribute Queries (5 questions):} Questions regarding object attributes like color, material, or location.
\end{itemize}

\begin{figure}[htbp]
\centering
\begin{minipage}[t]{0.58\linewidth}
  \vspace{0pt}
  \centering
  \begin{subfigure}[t]{0.48\linewidth}
    \includegraphics[width=\linewidth, height=2cm]{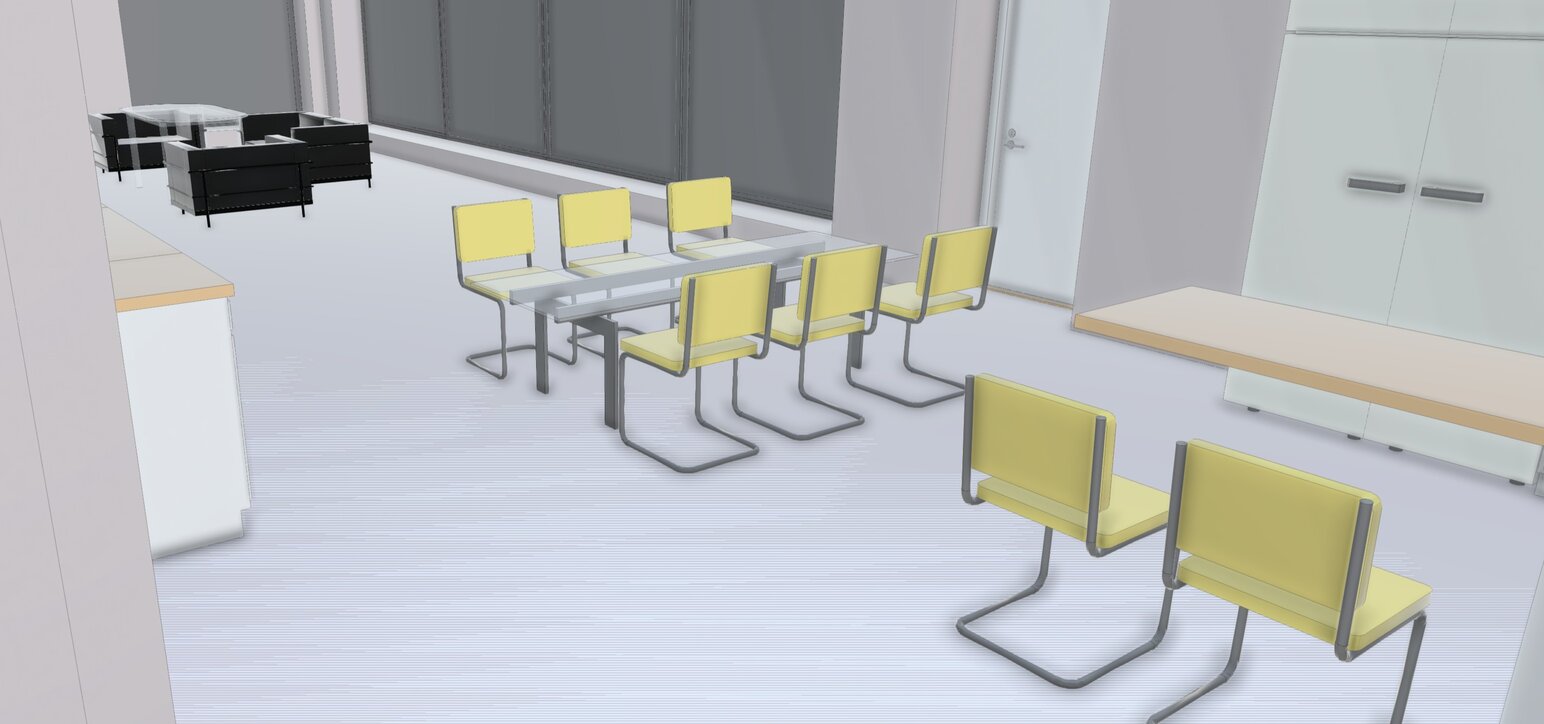}
  \end{subfigure}\hfill
  \begin{subfigure}[t]{0.48\linewidth}
    \includegraphics[width=\linewidth, height=2cm]{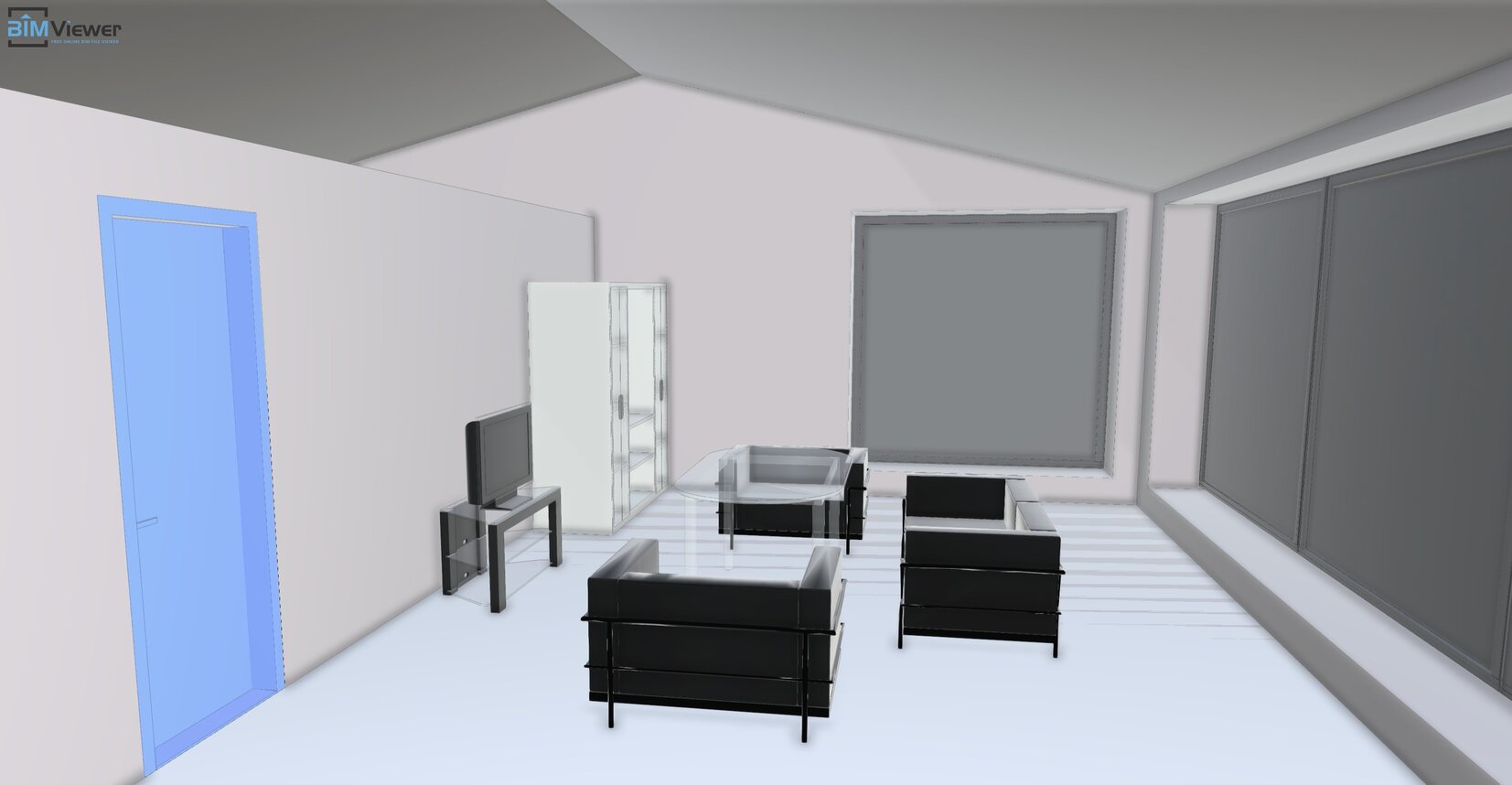}
  \end{subfigure}
  
  \vspace{2mm}
  
  \begin{subfigure}[t]{0.48\linewidth}
    \includegraphics[width=\linewidth, height=2cm]{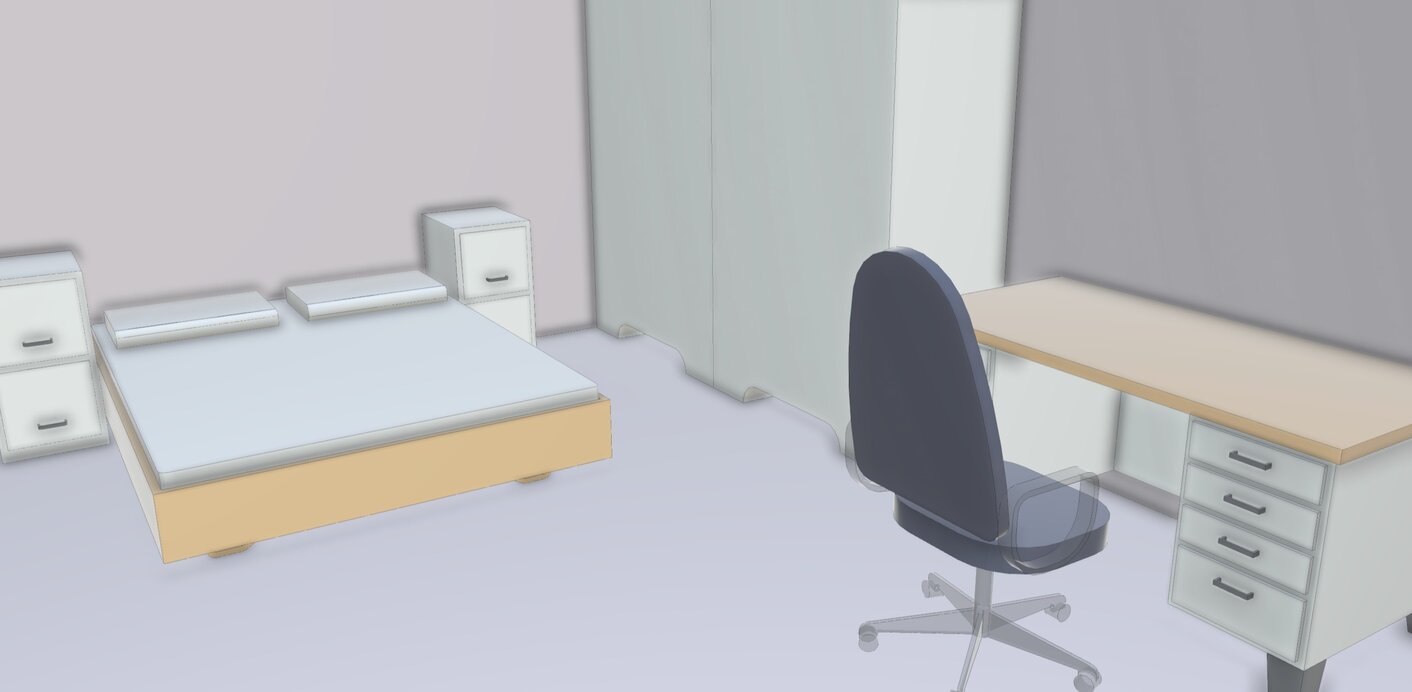}
  \end{subfigure}\hfill
  \begin{subfigure}[t]{0.48\linewidth}
    \includegraphics[width=\linewidth, height=2cm]{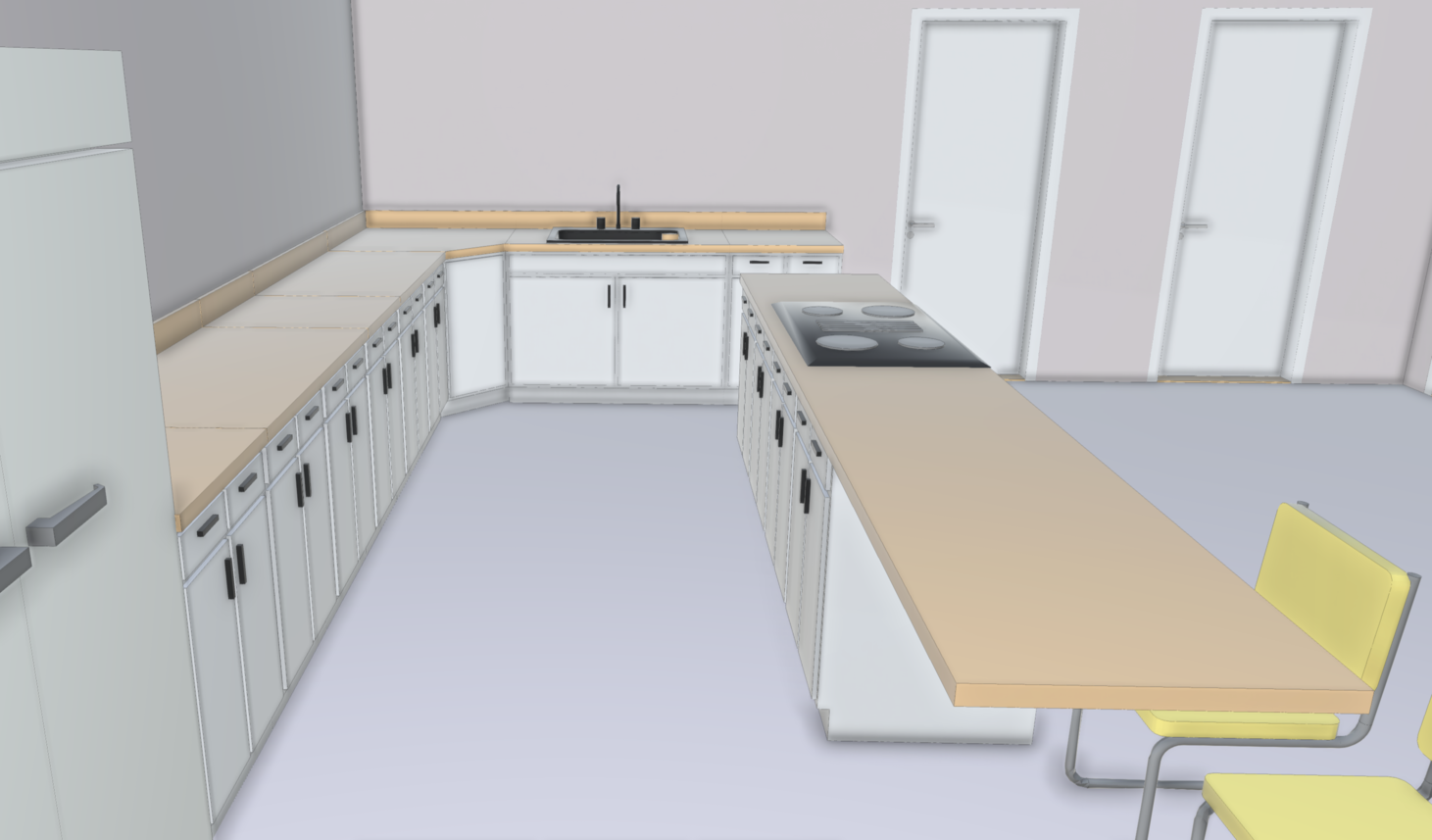}
  \end{subfigure}
\end{minipage}\hfill
\begin{minipage}[t]{0.40\linewidth}
  \vspace{0pt}
  \centering
  \begin{subfigure}[t]{0.48\linewidth}
    \includegraphics[width=\linewidth, height=4.2cm]{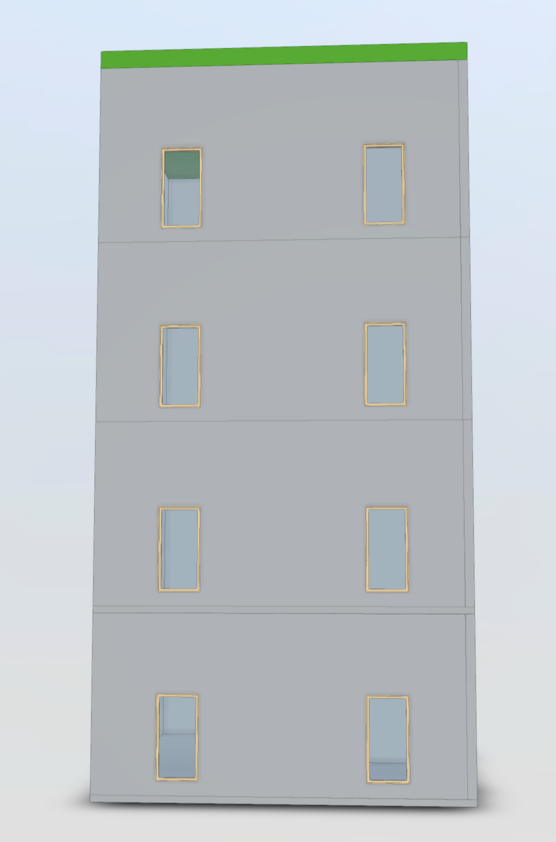}
  \end{subfigure}\hfill
  \begin{subfigure}[t]{0.48\linewidth}
    \includegraphics[width=\linewidth, height=4.2cm]{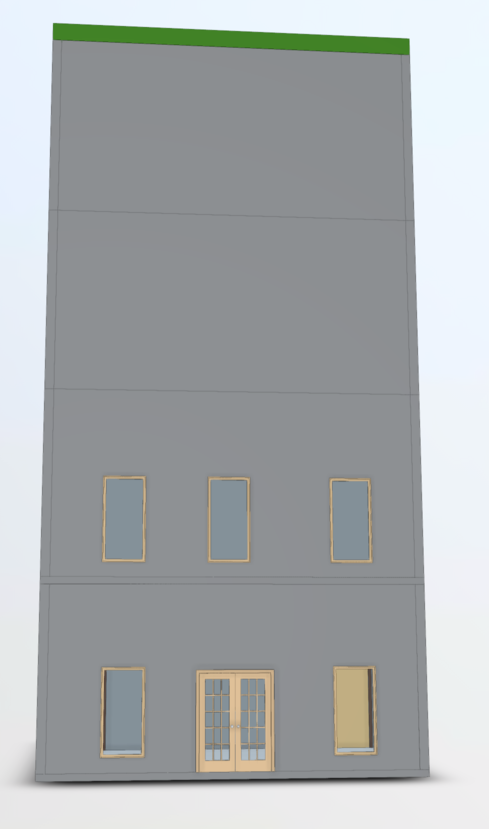}
  \end{subfigure}
\end{minipage}
\caption{Screenshots used for the visual question answering experiment. The first four images are shown in a 2×2 layout (top-left: P1, top-right: P2, bottom-left: P3, bottom-right: P4), followed by P5 and P6).}
\label{qa_images}
\end{figure}

In total, 40 questions were evaluated. Table~\ref{spatial_questions} shows a few examples of the image-based questions, and the corresponding screenshots are presented in Figure \ref{qa_images}. For evaluation, we used a three-level scoring system: 1.0 for correct answers, 0.5 for partially correct answers, and 0 for incorrect answers. Using this metric, Claude Sonnet 4.5 achieved an overall score of $73\%$. The model was often able to recognize major objects and answer attribute questions correctly. However, its performance on counting was less consistent, and some spatial relationships were incorrect. These results suggest that while the model can extract high-level visual information from BIM images, its understanding of scene geometry and the quantity of objects remains limited. Nonetheless, this simple experiment shows that screenshot-based visual feedback can already support certain interactions, and motivates stronger vision-language integration for future BIM applications. 

\subsubsection{Comparative Model Analysis}
\begin{figure}[t]
\centering
\includegraphics[width=0.8\columnwidth]{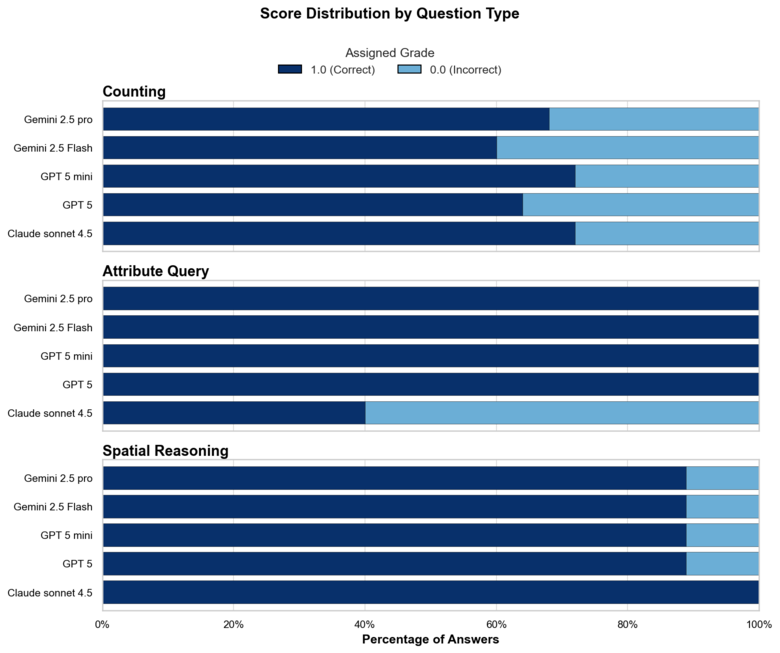}
\caption{Results of the Spatial Benchmark}
\label{fig:SpQA-Results}
\end{figure}

To better contextualize Claude Sonnet 4.5 performance, we compared its performance against four other multimodal models (GPT 5, GPT 5 mini, Gemini 2.5 Flash, and Gemini 2.5 pro) on the same 40-question dataset. While the overall scores were closely clustered, a categorical breakdown revealed distinct and contrasting performance profiles for each model, as shown in Figure \ref{fig:SpQA-Results}.
%
%
This comparative analysis suggests that no single model currently excels at all forms of visual reasoning for BIM. 
Claude Sonnet 4.5, for instance, demonstrates a better understanding of spatial geometry but struggles with simple attribute identification, a task other models find trivial. 
However, as these observations are curently based on a limited set of experiments and a single seed run, they should therefore be interpreted as preliminary trends rather than definitive conclusions about model suitability for specific BIM tasks.

\begin{table}[tb]
\centering
\small
\setlength{\tabcolsep}{4pt}
\renewcommand{\arraystretch}{1.15}
\caption{A few example questions for different building scenes.}
{
\rowcolors{2}{gray!5}{white}
\begin{tabular}{@{}> {\centering\arraybackslash}p{0.12\linewidth}p{0.83\linewidth}@{}}
\toprule
\textbf{Picture No.} & \textbf{Spatial Question} \\
\midrule
P1 & How many chairs are there in the image? \\
P1 & Where can you find a pair of chairs in the image? \\
P1 & What color are the chairs?  \\
\midrule
P2 & Where is the TV located relative to the sofa? \\
P2 & Is there a table in the room? \\
P2 & What element in the scene is blue? \\
\midrule
P3 & How many pillows are there on the bed? \\
P3 & What is in front of the chair? \\
P3 & How many drawers does the table have? \\
\midrule
P4 & Where is the sink in relation to the stove? \\
P4 & How many stoves are on the kitchen counter? \\
P4 & How many doors do you count in the kitchen? Give an exact number. \\
\midrule
P5/P6 & How many floors does this building have? \\
P5/P6 & How many windows are on this building? \\
P5/P6 & How many doors are on this building? \\
\bottomrule
\end{tabular}
}
\label{spatial_questions}
\end{table}

\end{document}

%% file: math_commands.tex
\usepackage{amsmath,amsfonts,bm}

\def\eqref#1{equation~\ref{#1}}

\def\1{\bm{1}}

\DeclareMathAlphabet{\mathsfit}{\encodingdefault}{\sfdefault}{m}{sl}
\SetMathAlphabet{\mathsfit}{bold}{\encodingdefault}{\sfdefault}{bx}{n}

